\documentclass[preprint,3p,5pt,authoryear]{elsarticle}

\usepackage{amssymb}
\usepackage{float}
\usepackage{amsmath}
\usepackage{algorithm}
\usepackage{algorithmic}
\usepackage{multirow}
\usepackage{longtable}
\usepackage{graphicx}
\usepackage{makecell}
\usepackage{xcolor}
\usepackage{booktabs}
\usepackage[colorlinks=true, allcolors=blue]{hyperref}

\journal{Engineering Applications of Artificial Intelligence}

\begin{document}
	
\begin{frontmatter}
	


\title{Unraveling the Rainbow: can value-based methods schedule?} 


\author[UCAddress]{Arthur Corrêa\corref{cor1}}
\author[UCAddress]{Alexandre Jesus}
\author[UCAddress]{Cristóvão Silva}
\author[UCAddress]{Paulo Nascimento}
\author[UCAddress]{Samuel Moniz}

\cortext[cor1]{Corresponding author.\\
	Email: ajpcorrea@dem.uc.pt\\
	Address: Pólo II da Universidade de Coimbra, Rua Luís Reis Santos, 3030-788, Coimbra, Portugal\\
	Phone: +351 914 704 871}

\affiliation[UCAddress]{organization={Department of Mechanical Engineering, CEMMPRE, ARISE, Universidade de Coimbra}, 
	city={Coimbra},
	country={Portugal}}

\begin{abstract}
	In this work, we conduct an extensive empirical study of several deep reinforcement learning algorithms on two challenging combinatorial optimization problems: the job-shop and flexible job-shop scheduling problems, both fundamental challenges with multiple industrial applications. Broadly, deep reinforcement learning algorithms fall into two categories: policy-gradient and value-based. While value-based algorithms have achieved notable success in domains such as the Arcade Learning Environment, the combinatorial optimization community has predominantly favored policy-gradient algorithms, often overlooking the potential of value-based alternatives. From our results, value-based algorithms demonstrated a lower variance and a more stable convergence profile compared to policy-gradient ones. Moreover, they achieved superior cross-size and cross-distribution generalization, that is, effectively solving instances that are substantially larger or structurally distinct from those seen during training. Finally, our analysis also suggests that the relative performance of each category of algorithms may be dependent on structural properties of the problem, such as problem flexibility and instance size. Overall, our findings challenge the prevailing assumption that policy-gradient algorithms are inherently superior for combinatorial optimization. We show instead that value-based algorithms can match or even surpass the performance of policy-gradient algorithms, suggesting that they deserve greater attention from the combinatorial optimization community. Our code is openly available at: \url{https://github.com/AJ-Correa/Unraveling-the-Rainbow/tree/master}
\end{abstract}


	
\end{frontmatter}



\section{Introduction}
\label{sec:intro}

Efficient scheduling of operations is vital across various industries, including semiconductor manufacturing \citep{Park2021} and aerospace \citep{Tian2023}, enabling higher resource utilization and lower operational costs. In this regard, the classical job-shop scheduling problem (JSSP) holds significant importance in both real-world industry applications and academia, being a well-known problem in the combinatorial optimization (CO) literature. The JSSP entails determining the optimal sequence of operations (according to a given objective function), where each operation must be processed on a predetermined machine. The flexible job-shop scheduling problem (FJSP) is an extension to the JSSP and was first called job-shop with multi-purpose machines \citep{Brucker1990}. Its complexity is further enhanced by allowing process flexibility, meaning that operations can be executed on any machine from the eligible set, instead of a predetermined machine. The assignment of operations to machines is called job-routing, and introduces an additional decision level to be considered, making the FJSP more challenging to solve than the JSSP.

Historically, CO problems have been predominantly solved by exact formulations and meta-heuristics \citep{Xiong2022,Zhang2022}. While exact algorithms guarantee optimality with sufficient computational time, they tend to become intractable when solving medium and large-sized instances \citep{Gomes2005,Ku2016}. Alternatively, meta-heuristics employ efficient exploration mechanisms to find high-quality solutions within reduced computational time \citep{Li2016,Pezzella2008AGA}. Still, these methods face significant limitations when applied to large-scale problems due to their non-linear time complexity. This challenge becomes even more pronounced when dealing with applications where problems must be solved in real-time.

In response, deep reinforcement learning (DRL) has emerged in recent years as an alternative solving method to CO problems \citep{Bengio2021}. By leveraging neural networks, DRL methods learn decision-making patterns through interactions with a simulated environment, solving these problems in an end-to-end manner without the need for handcrafted heuristics. Generally, DRL algorithms are classified into two main categories: value-based and policy-gradient methods. Value-based approaches focus on estimating the value of actions in a given state of the environment, where the state reflects the current situation of the problem. The goal is to learn which actions maximize long-term rewards, using the learned value function to indirectly derive a policy. On the other hand, policy-gradient methods directly optimize a policy that maps states to actions, without relying on an explicit value estimation. While policy-gradient methods often excel in environments with high-dimensional action spaces, value-based approaches are often more sample efficient and exhibit lower variance.

Among existing DRL algorithms, many value-based methods have demonstrated impressive performance, from which Rainbow has gained particular importance \citep{Hessel2017RainbowCI}. Rainbow combines six different improvements to the Deep Q-Network (DQN) algorithm \citep{Mnih2015}, resulting in a highly effective agent. However, most performance benchmarks used in DRL literature – especially those highlighting Rainbow – are based on Arcade Learning Environments (ALEs) \citep{Bellemare2013}, leaving the effectiveness of value-based methods largely unexplored outside such environments. In contrast, existing research on CO has focused primarily on policy-gradient methods \citep{Mazyavkina2021}, such as Proximal Policy Optimization (PPO) \citep{Schulman2017}. Consequently, there is a fundamental knowledge gap regarding whether value-based algorithms can effectively tackle classical NP-hard optimization problems. Moreover, they often demand significant implementation effort and computational resources \citep{Clark2025beyond}, leaving decision-makers uncertain about whether the potential benefits justify the added complexity and cost.

Motivated by this gap, we undertake an empirical study on the performance of several value-based algorithms on two complex CO problems: the JSSP and FJSP. This leads us to our main research question, which guides our entire investigation:

\vspace{0.5em}

1) \textit{Can value-based DRL algorithms compete with policy-gradient methods in job-shop scheduling problems?}

\vspace{0.5em}

Policy-gradient approaches currently dominate existing DRL-based approaches for the JSSP and FJSP, but it remains unclear whether this prevalence is justified performance-wise. To rigorously evaluate this, we first need to understand how value-based methods perform on their own. Consequently, another question that deserves to be addressed is:

\vspace{0.5em}

2) \textit{Do value-based algorithms, like Rainbow and its individual components, offer significant improvements over the DQN algorithm when solving job-shop scheduling problems?}

\vspace{0.5em}

The DQN algorithm is one of the most widely used and effective value-based DRL algorithms. Rainbow, in turn, integrates several enhancements that were introduced to improve upon the original DQN framework. Since Rainbow combines multiple extensions into a single architecture, we can evaluate the individual contribution of each component to the overall performance. Building on this idea, we aim to explore if the process flexibility, present in the FJSP, has any impact on the performance of these algorithms. This leads to the next question:

\vspace{0.5em}

3)	\textit{How does the performance of value-based methods vary when applied to problems with higher routing flexibility, such as the FJSP, compared to those with a fixed routing structure, like the JSSP?}

\vspace{0.5em}

This allows us to assess whether the results observed in research question 2 remain consistent when the problem structure changes, and whether certain algorithmic enhancements are better suited to flexible scheduling scenarios. Finally, understanding an algorithm’s performance on training instances is not sufficient. We must also assess how well these methods generalize beyond their original training setting. This gives rise to our final research question:

\vspace{0.5em}

4)	\textit{How well do different value-based algorithms generalize across instances with varying sizes and distributions?}

\vspace{0.5em}

To answer the aforementioned questions, we designed a series of controlled experiments, including a detailed ablation study of Rainbow’s individual extensions over the DQN algorithm. This analysis helps to isolate the effect of each component on the performance of the model. As a result, we can understand how each value-based extension performs in terms of numerous factors, such as instance size, problem flexibility, generalization (beyond the original training problem’s size and distribution), sample efficiency, stability and convergence speed. Moreover, by measuring the effectiveness of each extension, we can hypothesize which theoretical advantages explain the observed behaviors, allowing us to recommend which extensions are better suited for each scenario.

Then, to situate value-based approaches within the broader DRL landscape, we assess their effectiveness against a series of state-of-the-art policy-gradient methods, including REINFORCE \citep{Williams1992}, Advantage Actor-Critic \citep{Mnih2016}, PPO \citep{Schulman2017}, and On-policy Maximum a Posteriori Policy Optimization \citep{Song2020V-MPO}.

The main contributions of our work are as follows:
\begin{itemize}
	\item Extensive comparison of policy-gradient and value-based methods in scheduling problems: To the best of our knowledge, our work provides the first extensive empirical comparison between policy-gradient and value-based algorithms on scheduling problems. This contribution addresses a critical gap in the literature, where the dominance of policy-gradient approaches has not been thoroughly investigated and discussed. We argue that value-based methods deserve more attention due to various theoretical advantages, such as better sample efficiency and more stable convergence, which may translate into competitive performance in practice. Our analysis provides important insights into the strengths and limitations of each category of methods, helping decision-makers, practitioners and researchers select the most appropriate learning strategies depending on the problem setting.
	\item Novel empirical results across different problems and instance sizes: Extensive results on public benchmark and randomly generated instances show that, on numerous occasions, value-based methods perform on par, or even better than policy-gradient. Our analysis further shows that value-based algorithms offer several distinct advantages over policy-gradient. For instance, they displayed a lower variance during training, leading to a more stable convergence profile. They also achieved better generalization to larger instances and instances drawn from parameter distributions different from those seen during training. In addition, we observed that the Distributional RL algorithm was effective in dealing with the hard credit assignment challenge of the JSSP, while the Multi-step learning extension performed well across both JSSP and FJSP. Lastly, we observed that the effectiveness of both categories of methods is strongly influenced by structural properties of the problem, such as the flexibility in machine assignment decisions, and the size of the instance solved.
	\item End-to-end DRL environment supporting customizable Rainbow configurations: We developed a flexible DRL experimentation environment, centered around a graph neural network (GNN) model, which supports numerous value-based and policy-gradient algorithms. The environment allows to activate/deactivate each Rainbow component, enabling up to 64 distinct algorithmic configurations (apart from hyperparameters). This modular design facilitates systematic comparisons, ablation studies, and robust benchmarking of DRL algorithms in the scheduling literature. Our code is openly available at: \url{https://github.com/AJ-Correa/Unraveling-the-Rainbow/tree/master}
\end{itemize}

The remainder of this paper is organized as follows. Section~\ref{sec:review} discusses the related work in JSSP and FJSP literature. Section~\ref{sec:prelim} presents important preliminaries to our work, including the problems’ formulations and a technical explanation of all DRL algorithms employed in this study. Section~\ref{sec:architecture} describes all architectural details of our model. Section~\ref{sec:experiments} presents the computational experiments and discusses the findings. Finally, Section~\ref{sec:conclusion} concludes the paper and proposes future research directions. 

\section{Related work}
\label{sec:review}

\begin{table}[h]
	\label{table_drl_review}
	\centering
	\caption{DRL methods for job-shop scheduling problems}
	\renewcommand{\arraystretch}{1.15}
	\begin{tabular}{l c c c c}
		\toprule
		\textbf{Method} & \textbf{Year} & \textbf{Problem} & \textbf{Category} & \textbf{Algorithm} \\
		\midrule
		\citet{Zhang2020} & 2020 & JSSP & Policy-gradient & PPO \\
		\citet{Han2020} & 2020 & JSSP & Value-based & Dueling Double DQN \\
		\citet{Park2020} & 2020 & FJSP-SDST & Value-based & DQN \\
		\citet{Park2021} & 2021 & JSSP & Policy-gradient & PPO \\
		\citet{Tassel2021} & 2021 & JSSP & Policy-gradient & PPO \\
		\citet{Wang2021} & 2021 & DyJSSP & Policy-gradient & PPO \\
		\citet{Luo2021} & 2021 & DyFJSP & Value-based     & Double DQN \\
		\citet{Lei2022} & 2022 & FJSP & Policy-gradient & PPO \\
		\citet{Song2023} & 2023 & FJSP & Policy-gradient & PPO \\
		\citet{Tassel2023} & 2023 & JSSP & Policy-gradient & Imitation learning + Policy-gradient \\
		\citet{Zhang2023deepmag} & 2023 & FJSP        & Value-based     & DQN \\
		\citet{Iklassov2023} & 2023 & JSSP & Policy-gradient & REINFORCE \\
		\citet{Yuan2024} & 2024 & FJSP & Policy-gradient & PPO \\
		\citet{Wang2024} & 2024 & FJSP & Policy-gradient & PPO \\
		\citet{Ho2024} & 2024 & JSSP/FJSP & Policy-gradient & REINFORCE \\
		\citet{Jing2024} & 2024 & FJSP & Policy-gradient & DDPG \\
		\citet{Echeverria2025} & 2025 & FJSP & Policy-gradient & PPO \\
		\bottomrule
	\end{tabular}
\end{table}

Over recent years, a variety of DRL-based approaches have been developed to solve scheduling problems. By learning powerful data-driven heuristics from unlabeled data, these methods can generate high-quality schedules on large-scale instances in a few seconds. In this section, we review these studies, highlighting the challenges that remain open in the literature, and how they motivate our work.

\citet{Zhang2020} proposed Learning to dispatch (L2D), the first size-agnostic JSSP model in the literature. The authors used a graph isomorphism network (GIN) to encode the partial solution states through a disjunctive graph representation, and trained L2D using PPO, a classic policy-gradient algorithm. Overall, for a first attempt at a learning-based model, L2D showed good effectiveness, outperforming classic dispatching rules in randomly generated and benchmark instances. Around the same time, \citet{Han2020} investigated the use of a value-based algorithm for the JSSP. Instead of a GIN, they represented partial solution states with a deep convolutional neural network and trained the model using a Dueling Double DQN algorithm. Although their work was among the first to apply a value-based algorithm to scheduling, the absence of direct comparisons with policy-gradient or other value-based algorithms left the relative effectiveness of value-based DRL for the JSSP unclear. A related (though more system-oriented) line of work was presented by \citet{Park2020}, who applied a DQN-based approach to solve a semiconductor manufacturing problem, modeled as a FJSP with sequence-dependent setup times (FJSP-SDST). Their framework employs a multi-agent architecture where each machine is modeled as an independent agent that acts based on both local and global information. While exploring a value-based algorithm, unlike L2D, the proposed model is not size-agnostic. As the length of the state depends directly on the number of operations in the problem, the model must be retrained whenever the number of operations changes, which limits scalability for real-world applications. Additionally, like the approach of \citet{Han2020}, it does not conduct any direct comparison with policy-gradient methods.

As literature evolved, efforts to refine learning representation continued. Later, \citet{Park2021} introduced Learning to Schedule, a similar approach to L2D, also trained with PPO, that features a customized message passing architecture to account for different edge types in the graph structure. Other works moved in a different direction from graph-based encoders. \citet{Tassel2021} developed a DRL environment for the JSSP using a matrix-based state representation composed of various hand-crafted features. Although their method yielded promising results within a limited training budget, it lacked scalability, requiring retraining for problems with a different number of jobs. A similar departure from graph-based representations was introduced by \citet{Wang2021}, who solved the dynamic JSSP (DyJSSP) with a new Markov decision process (MDP) formulation, modeling the state representation with three different matrices: one for job processing status, one for machine states, and one for operation processing times. Like previous methods, the proposed approach was also trained with PPO. 

An additional line of research investigates hierarchical DRL, a class of DRL methods which decomposes the decision-making process into multiple levels of abstraction instead of relying on a single policy. \citet{Luo2021} built a multi-objective model trained with a double DQN algorithm for the DyJSSP. Their approach incorporates two different agents for (i) selecting which objective will be optimized, and (ii) constructing scheduling solutions, deciding the specific actions needed to achieve the subgoal chosen by the higher agent. Later, \citet{Lei2022} proposed a multi-action model for the FJSP, trained with PPO. In the proposed method, two distinct policies are learned: one to select which job to dispatch, and another to determine the machine on which it will be processed. The problem is modeled as a disjunctive graph, and the state is encoded using a GIN, making the proposed model size-agnostic.

A more recent contribution by \citet{Song2023} represents a significant step forward in FJSP research. The authors extended the traditional disjunctive graph into a heterogeneous representation, providing the model with more informative states through specific machine nodes. \citet{Tassel2023} proposed a novel perspective by leveraging a constraint programming solver to declaratively encode scheduling solutions using variables, constraints, and objective functions. \citet{Zhang2023deepmag} introduced DeepMAG, a multi-agent system with graph-based representations to model job and machine agents, trained using DQNs. Through coordination mechanisms, agents learn to cooperate with each other to optimize the overall performance of the system. A different strategy was explored by \citet{Iklassov2023}, who introduced a recurrent long short-term memory architecture, trained with a curriculum learning regimen to improve generalization by gradually increasing the difficulty of instances over training (measured by the number of jobs and machines in the problem). Their method demonstrated substantial performance gains, significantly outperforming L2D on different benchmark datasets. \citet{Yuan2024} reduced the computational complexity compared to existing methods by using a lightweight multi-layer perceptron (MLP) as the encoding network. They also designed a new action space that simultaneously considers both subproblems of the FJSP, i.e., machine assignment and operation sequencing.

Among more recent approaches, DANIEL \citep{Wang2024} represents a notable advancement, utilizing self-attention mechanisms to effectively capture operation precedence and machine flexibility dependencies. The model employs a dual-attention network architecture with interconnected attention blocks for operations and machines, enabling more expressive representations for the FJSP. Subsequent methods continued to explore ways to refine these representations. \citet{Ho2024} introduced Residual Scheduling, an approach designed for both the JSSP and FJSP. This method removes, at each timestep, irrelevant nodes from the partial schedule, such that only remaining and relevant machines and jobs are kept in the graph state. This way, the agent can direct its attention to relevant nodes that still have not been added to the solution. However, removing irrelevant nodes may inadvertently discard contextual information that could be useful for completing the schedule. \citet{Jing2024} take a different route by modeling the FJSP as a probabilistic directed acyclic graph. The authors adopted a multi-agent setup with multiple job and machine agents. To represent the scheduling environment, a graph convolutional network is employed to encode the graph structure. The scheduling policy is learned by predicting edge connection probabilities, framing the FJSP as a topological graph prediction process. \citet{Echeverria2025} extended the heterogeneous graph representation by introducing a new job-type node, improving the model’s ability to capture higher-level job structures within the graph. The authors also designed a mechanism which artificially reduces the action space at each timestep, excluding actions that create large gaps in the scheduling, and thus focusing the policy on more promising decisions. Additionally, the paper proposes a diverse policy generation framework that trains multiple distinct models. To identify the most effective policies, a K-Nearest Neighbors algorithm is applied for self-evaluation and selection. While achieving good performance on various benchmarks, the action-space reduction may occasionally prune high-quality solutions. A summary of the aforementioned DRL approaches for scheduling problems is provided in Table~\ref{table_drl_review}. 

The growing body of scheduling research has evolved from early size-agnostic policies to more sophisticated representations and learning strategies. Only more recently, \citet{Wang2024} conducted a comprehensive review of different DRL design patterns for the JSSP. The authors identified several patterns, including algorithm selection (e.g., policy-gradient vs. value-based methods), state representation strategies (such as graph- or matrix-based encodings), reward function design, action space formulation, and the overall solution construction process. Their retrospective analysis shows how the literature has diversified and provides a useful framework for evaluating and comparing future methods.

Addressing the relative lack of attention to value-based methods in scheduling, \citet{Correa2024} investigated the performance of Rainbow in comparison with DQN on the FJSP. Although this marks a step toward examining value-based approaches, their evaluation was limited in scope. Notably, the individual contributions of Rainbow’s algorithmic components were not separately evaluated. Moreover, the work did not include any comparison with policy-gradient approaches.

Despite these recent contributions, most research continues to emphasize policy-gradient methods, with PPO being particularly prominent. While some works have explored value-based algorithms, these approaches suffer from one or more of the following limitations. First, they lack size-agnostic properties, meaning that models trained on specific problem sizes (measured by the number of jobs, operations, and/or machines) cannot generalize to instances of different scales. Second, there is an absence of systematic comparisons with policy-gradient methods, which makes it difficult to understand how they perform compared to established policy algorithms. Third, the value-based algorithms that have been applied are relatively simple, even though more advanced methods like multi-step learning, noisy networks and Rainbow have shown superior performance in domains like ALEs. Consequently, the effectiveness of value-based algorithms on scheduling problems still remains quite uncertain. To the best of our knowledge, no prior work has systematically evaluated the performance of such algorithms on scheduling problems.

\section{Preliminaries}
\label{sec:prelim}

\subsection{Problem description}
\label{sec:prob_description}

\noindent\textbf{Job-shop scheduling problem:} A typical JSSP is defined as follows: two sets are given — $\mathcal{J} = \{J_1, J_2, \allowbreak \dots, \allowbreak J_n\}$ being a set of $n$ jobs and $\mathcal{M} = \{M_1, M_2, \allowbreak \dots, \allowbreak M_m\}$ a set of $m$ machines. An instance is characterized by dimension $n \times m$, indicating the number of jobs and machines, respectively. Each job $J_i \in \mathcal{J}$ consists of a sequence of $N_i$ operations $O_{i1} \xrightarrow{} O_{i2} \xrightarrow{} ... \xrightarrow{} O_{iN_i}$, where each operation $O_{ij}$ can only begin after the completion of the preceding operation $O_{i(j - 1)}$. For each operation $O_{ij}$, there is a corresponding machine $M_k$ responsible for processing it, with a processing time $p_{ij}$. The problem is subject to further constraints: once a machine starts to process an operation, it must complete it without interruption, and each machine can process only one operation at a time. Additionally, all jobs and machines are available at the beginning of the scheduling horizon. In this paper, we consider minimizing the makespan $C_{max}$ as the objective function, where $C_{max} = max_i\{C_i\}$, with $C_i$ being the completion time of job $J_i$ $\forall$ $i \in \{1, 2, ..., n\}$. In simple terms, the makespan is the total time necessary for completing all jobs.

\vspace{0.5em}

\begin{figure*}[h]
	\centering
	\includegraphics[width=\textwidth]{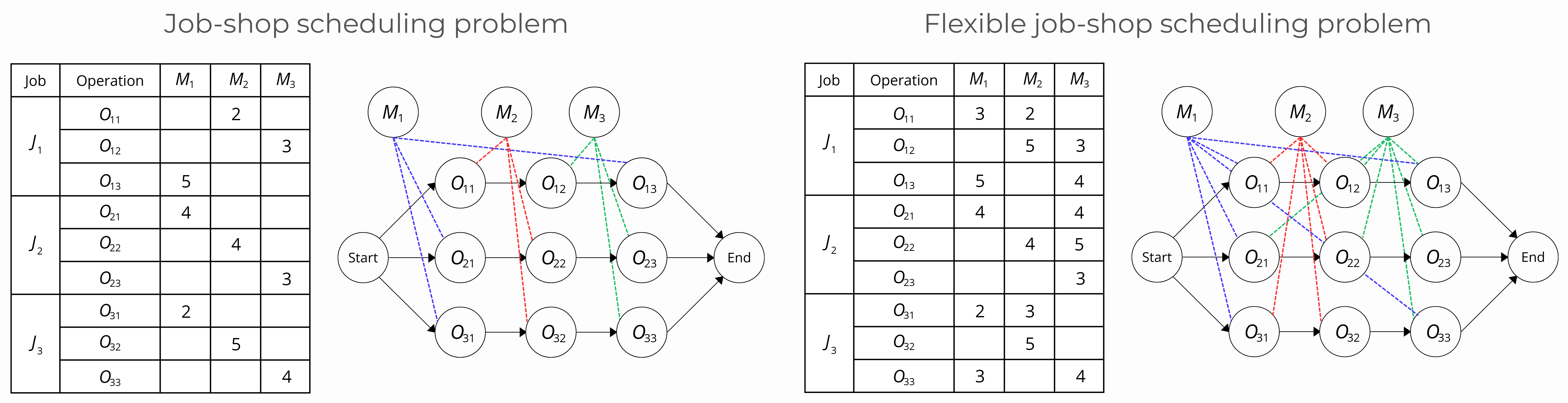}
	\caption{JSSP and FJSP instances.}
	\label{fig_jssp_instances}
\end{figure*}

\noindent\textbf{Flexible job-shop scheduling problem:} The FJSP is an extension of the JSSP that introduces flexibility in machine assignment decisions. In a standard JSSP, each operation must be processed on a predetermined machine. Instead, in the FJSP, each operation $O_{ij}$ can be processed by a subset of candidate machines $\mathcal{M}_{ij} \subseteq \mathcal{M}$. For each candidate machine $M_k \in \mathcal{M}_{ij}$, the processing time of $O_{ij}$ is denoted by $p_{ijk}$. This added flexibility decouples the strict one-to-one mapping between operations and machines. The FJSP has total flexibility if, for every operation, the set of candidate machines is equal to $\mathcal{M}$. In contrast, it has partial flexibility if at least one operation is not executable on all machines. Similarly to the JSSP, the objective function is to minimize the makespan. All other JSSP constraints also apply to the FJSP. Two examples of very simple 3x3 instances for the JSSP and FJSP are given in Figure~\ref{fig_jssp_instances}, with the processing times for each operation-machine pair on the respective tables.

\subsection{Reinforcement learning}
\label{sec:rl}

In reinforcement learning (RL) problems, one or more agents operate within a given environment and interact with it over a sequence of discrete timesteps. At each timestep $t$, the agent observes the environment state $s_t$ and selects an action $a_t$ from the set of possible actions $\mathcal{A}_t$ at that timestep. This action influences the environment, leading to a new state $s_{t+1}$ determined by a transition function $P$. Finally, the agent is given a reward $r_t$ based on a reward function $R(s_t, a_t)$. This loop continues until training is completed. Another important notion in RL is the definition of episodes. An episode is a complete sequence of interactions, starting from an initial state and ending upon reaching a terminal state. It corresponds to one full attempt by the agent to accomplish the given task. In a JSSP/FJSP instance, for example, an episode encompasses the entire construction of a schedule. At each timestep, the action consists of selecting an eligible operation-machine pair, and the episode terminates once all operations have been assigned.

The aforementioned interactions can be formalized as an MDP \citep{Puterman1994}, which provides a mathematical framework for sequential decision-making tasks. An MDP is defined by a tuple $(\mathcal{S}, \mathcal{A}, R, P, \gamma)$, where: $\mathcal{S}$ denotes the set of states, $\mathcal{A}$ represents the action space, $R$ is the reward function, $P$ denotes the transition function and $\gamma$ is the discount factor.

The goal of an RL agent is to maximize its total cumulative rewards, also known as return. This return can be expressed as $G_t = \sum_t \gamma^t r_t$, where $\gamma \in [0, 1)$ is a discount factor that balances the significance of short-term and long-term rewards. The agent learns a policy $\pi$ to maximize its expected return. Therefore, it is crucial that the reward function aligns with the objective function, ensuring that maximizing long-term rewards leads to optimal solutions. Building on this foundation, \textit{value-based and policy-gradient methods differ mainly in what they learn and how they represent decision-making}. In value-based RL, the policy is obtained by first estimating either: (i) a state-value function $V_\pi(s) = \mathbb{E}_\pi[G_t|s_t = s]$, which gives the expected discounted return starting from state $s$ and following the policy onward, or (ii) an action-value (or q-value) function $Q_\pi(s, a) = \mathbb{E}_\pi[G_t|s_t = s, a_t = a]$, which also considers the action taken at step $t$. The policy is then derived by selecting actions that maximize the appropriate value estimate. In contrast, policy-gradient algorithms learn the policy directly. They adjust the policy to increase the likelihood of actions that lead to higher rewards, producing an explicit (often stochastic) mapping from states to actions, without requiring an intermediate value function. 

\subsection{Policy-gradient DRL}
\label{sec:pg_drl}

As explained in the previous subsection, policy-gradient methods estimate a stochastic policy $\pi_{\theta}(a_t \ | s_t)$, where $\pi_{\theta}$ denotes the probability of taking action $a_t$ given a state $s_t$, parametrized by $\theta$. The objective is to find the set of parameters $\theta$ that maximize the objective function $J(\theta) = \mathbb{E}_{\pi_{\theta}}[\sum_{t=0}^{T} \ \gamma^{t} r_t]$, which measures the expected return over trajectories generated by the policy. Here, a trajectory is a sequence of transitions $(s_0, a_0, r_0, s_1, a_1, r_1, ..., s_T, a_T, r_T)$ that composes an episode, with $T$ being the number of steps needed to reach the terminal state. The parameters $\theta$ are then optimized via gradient ascent. Ahead we present the policy-gradient methods considered in our experimental setting.

\noindent\textbf{REINFORCE:} \citet{Williams1992} introduced REINFORCE, one of the earliest and most fundamental policy-gradient algorithms in the literature. In this method, the gradient of the objective function is expressed as:
\begin{equation}
	\nabla_{\theta} J(\theta) = \mathbb{E}_{\tau \sim \pi_{\theta}} \left[\sum_{t=0}^T \nabla_{\theta} \log \pi_{\theta}(a_t|s_t) G_t \right],
\end{equation}
where $\tau$ is a trajectory sampled from $\pi_{\theta}$.

At each iteration, the algorithm collects a sample of trajectories using the current policy $\pi_{\theta}$, computes the return $G_t$ for each timestep $t$, and updates the policy parameters via stochastic gradient ascent:
\begin{equation}
	\theta \leftarrow \theta + \eta \nabla_{\theta} \log \pi_{\theta}(a_t|s_t) G_t,
\end{equation}
where $\eta$ is the learning rate.

Although REINFORCE is conceptually simple and easy to implement, it suffers from high variance, especially in long-horizon and sparse reward environments, which may make the learning process unstable. Nonetheless, REINFORCE remains a foundational method, inspiring many other works in the CO literature \citep{Kool2019,Kwon2020,Iklassov2023,Ho2024}.

\vspace{0.5em}

\noindent\textbf{Advantage Actor-Critic:} The Advantage Actor-Critic (A2C) algorithm \citep{Mnih2016} improves upon REINFORCE by combining elements from policy (actor) and value (critic) methods to reduce the variance of gradient estimates and improve sample efficiency. In A2C, the actor collects trajectories from the environment using the current policy $\pi_{\theta}$, whereas the critic learns a state-value function $V(s_t)$. This state-value function is used to compute the advantage of taking an action $a_t$ at state $s_t$:
\begin{equation}
	A_t = G_t - V(s_t),
\end{equation}
which quantifies how much better or worse an action is compared to the expected return from the state $s_t$. The policy gradient is then updated as:
\begin{equation}
	\nabla_{\theta} J(\theta) = \mathbb{E} \left[ \nabla_{\theta} \log \pi_{\theta}(a_t|s_t) A_t \right],
\end{equation}
with the policy parameters updated as $\theta \leftarrow \theta + \eta \nabla_{\theta} J(\theta)$.

In parallel, the critic updates its parameters by minimizing the mean squared error between its value estimates and the empirical returns:
\begin{equation}\label{eq_critic_loss}
	\mathcal{L}_{\text{critic}} = \mathbb{E}[(G_t - V(s_t))^2].
\end{equation}

\vspace{0.5em}

\noindent\textbf{Proximal policy optimization:} The PPO algorithm \citep{Schulman2017} is a widely used policy-gradient algorithm in the scheduling literature. It has better training stability and sample efficiency compared to previous methods by avoiding overly large policy updates. It is an actor-critic algorithm that introduces a surrogate objective with a clipping mechanism to constrain policy changes within a small range.

Let $\rho_t(\theta)$ denote the probability ratio between the current and old policies, $\pi_{\theta}$ and $\pi_{\theta_{\text{old}}}$ respectively:
\begin{equation}
	\rho_t(\theta) = \frac{\pi_{\theta}(a_t \ | \ s_t)}{\pi_{\theta_{\text{old}}}(a_t \ | \ s_t)}.
\end{equation}

The clipped surrogate objective function is defined as follows:
\begin{equation}
	\mathcal{L}^{\text{CLIP}}(\theta) = \mathbb{E}[\text{min}(r_t(\theta)A_t, \text{clip}(r_t(\theta), 1 - \delta, 1 + \delta)A_t)],
\end{equation}
where $\delta$ is a small positive hyperparameter that controls the maximum allowed deviation from the old policy.

The PPO algorithm also includes a state-value function loss (which refers to the Critic, same as in Equation~\ref{eq_critic_loss}), and an entropy bonus to encourage exploration.

\vspace{0.5em}

\noindent\textbf{On-policy maximum a posteriori policy optimization:} \citet{Song2020V-MPO} introduced on-policy maximum a posteriori policy optimization (V-MPO), a policy-gradient method designed to address the instability of large policy updates in high-dimensional action spaces. Unlike standard policy-gradient algorithms, which directly optimize the expected return, V-MPO first constructs a target distribution for the policy update by weighing the actions according to their estimated advantage function values. The policy is then updated with gradients that move its parameters toward this target.

The V-MPO algorithm introduces a new \textit{policy improvement} loss function, which is the sum of three components. The first one is the policy loss:
\begin{equation}
	\mathcal{L}_{\pi}(\theta) = - \sum_{s, a \sim \tilde{\mathcal{D}}} \psi(s, a) \ \text{log}\pi_{\theta}(a \ | \ s),
\end{equation}
where $\tilde{\mathcal{D}}$ are samples corresponding to the top 50\% of advantages and  $\psi(s, a)$ is a normalized weight over state-action pairs, computed to focus on actions with a higher advantage. More specifically, $\psi(s, a)$ is defined as:
\begin{equation}
	\psi(s, a) = \frac{\text{exp}(\frac{A^{\text{target}}(s, a)}{\eta_{\text{V-MPO}}})}{\sum_{s, a \sim \tilde{\mathcal{D}}}\text{exp}(\frac{A^{\text{target}}(s, a)}{\eta_{\text{V-MPO}}})},
\end{equation}
where $\eta_{\text{V-MPO}}$ is a Lagrange multiplier that acts as a temperature parameter to control the sharpness of the weighting.

The second component refers to the "temperature" loss, defined as:
\begin{align}
	\mathcal{L}_{\eta_{\text{V-MPO}}}(\eta_{\text{V-MPO}}) =\ & \eta_{\text{V-MPO}} \ \epsilon_{\eta_{\text{V-MPO}}} + \eta_{\text{V-MPO}} \ \text{log}\left[\frac{1}{|\tilde{\mathcal{D}}|}\sum_{s,a \sim \tilde{\mathcal{D}}} \text{exp}\left(\frac{A^{\text{target}}(s, a)}{\eta_{\text{V-MPO}}}\right)\right].
\end{align}

Here, $\epsilon_{\eta_{\text{V-MPO}}}$ is a Kullback-Leibler (KL) divergence bound and $\eta_{\text{V-MPO}}$ is optimized to enforce a soft KL constraint on how much the policy is allowed to change.

The third component enforces a trust region by limiting the deviation between the updated policy $\pi_{\theta}$ and a target policy $\pi_{\theta_{\text{target}}}$. More specifically, it computes the KL divergence between both policies and penalizes the updated policy if it exceeds a predefined threshold $\epsilon_{\alpha_{\text{V-MPO}}}$. The loss function term is defined as:
\begin{multline}
	\mathcal{L}_{\alpha} (\theta, \alpha) = \frac{1}{\mathcal{D}} \sum_{s \in \mathcal{D}} \big[
	\text{sg} \left[[ \alpha_{\text{V-MPO}} \right]] D_{\text{KL}}\left( \pi_{\theta_{\text{target}}}(a \mid s) \,\|\, \pi_{\theta}(a \mid s) \right) \\
	+ \alpha_{\text{V-MPO}} \left( \epsilon_{\alpha_{\text{V-MPO}}} - \text{sg} \left[[ D_{\text{KL}}\left( \pi_{\theta_{\text{target}}}(a \mid s) \,\|\, \pi_{\theta}(a \mid s) \right) \right]] \right) \big],
\end{multline}
where $D_{\text{KL}}$ is the KL divergence measuring the difference between both policies. The variable $\alpha_{\text{V-MPO}}$ dynamically adjusts the strength of the KL divergence constraint, and the stop-gradient (sg) prevents gradients from flowing through specific terms. In addition, the V-MPO algorithm includes a state-value function loss term, (a critic loss analogous to Equation~\ref{eq_critic_loss}).

\subsection{Deep q-network}
\label{sec:dqn}

The DQN algorithm \citep{Mnih2015} combines q-learning, RL and deep neural networks to handle high-dimensional state-action spaces. Unlike traditional q-learning \citep{watkins1992}, which uses a table to update the q-value of each state-action pair (intractable in large or continuous state spaces), DQN approximates the action-value function through neural networks. DQN learns by minimizing the loss between expected and target q-values, computed with:
\begin{equation}
	\mathcal{L}_{\text{DQN}} = \mathbb{E}_{(s_t, a_t, r_t, s_{t+1}) \sim \mathcal{D}} [(r_t + \gamma max_{a_{t+1}}Q_{\theta^{-}}(s_{t+1}, a_{t+1}) - Q_{\theta}(s_{t}, a_{t}))^2],
\end{equation}
where $\theta$ denotes the parameters (weights) of the neural networks, which are updated via backpropagation, and $\mathcal{D}$ is an experience buffer of transitions. To stabilize learning, DQN uses two separate networks: the online network $(\theta)$ and the target network $(\theta^{-})$. The online network is responsible for selecting actions and estimating current q-values, and its parameters are updated at every training step. The target network, on the other hand, is used to compute the target q-values, corresponding to the highest q-value on the next state. Its weights are updated less frequently by copying the parameters from the online network at regular intervals, which improves training stability by keeping the targets fixed over short periods.

Another key component of the DQN algorithm lies in the use of an experience buffer $\mathcal{D}$. At each timestep $t$, the transition tuple $(s_t,a_t,r_t,s_{t+1})$ is stored inside an experience replay buffer. When computing the loss function, a minibatch of transitions is uniformly sampled from the buffer to mitigate bias introduced by the temporal correlation between consecutive timesteps.

Finally, to balance exploration and exploitation, an $\varepsilon$-greedy strategy is employed. At each timestep, a parameter $\varepsilon$ ranging from 0 to 1 dictates the probability of selecting a random action rather than the one with the highest q-value. Typically, $\varepsilon$ is annealed over time, prioritizing random actions early on and gradually shifting toward exploitation as learning progresses.

\subsection{Rainbow}
\label{sec:rainbow}

\citet{Hessel2017RainbowCI} introduced Rainbow, an agent that combines six well-known algorithmic extensions to the original DQN. All extensions, individually and in combination, demonstrated significant improvements over the DQN algorithm when applied on the ALE. Next, we provide a brief overview of these extensions and refer the readers to the original papers for a more detailed description of each.

\vspace{0.5em}

\noindent\textbf{Double deep q-network:} In the original DQN, there is an overestimation bias since the same function is used for both action selection and evaluation when computing the target q-values. Double DQN (DDQN) \citep{Hasselt2016} mitigates this issue by decoupling action selection from evaluation: the online network selects the action $a_{t+1} = \text{arg max}_{a'} Q_{\theta}(s_{t+1}, a')$, while the target network evaluates its value $Q_{\theta^{-}}(s_{t+1}, a_{t+1})$. The corresponding loss is calculated as:
\begin{equation}
	\mathcal{L}_{\text{DDQN}} = \mathbb{E}_{(s_t, a_t, r_t, s_{t+1}) \sim \mathcal{D}} [(r_t + \gamma Q_{\theta^{-}}(s_{t+1}, \text{arg max}_{a'} Q_{\theta}(s_{t+1}, a')) - Q_{\theta}(s_{t}, a_{t}))^2].
\end{equation}

\vspace{0.5em}

\noindent\textbf{Prioritized Experience Replay:} Instead of sampling transitions uniformly, prioritized experience replay (PER) \citep{Schaul2016} assigns a higher sampling probability to transitions with greater temporal-difference errors, i.e., transitions with greater learning potential.

\vspace{0.5em}

\noindent\textbf{Dueling Networks:} \citet{Wang2016} proposed dueling networks, which modify the original DQN architecture by splitting the network into two separate streams: one for estimating the state-value function $V_\pi(s)$ and other for computing the advantage function $A_{\pi}(s, a) = Q_{\pi}(s, a) - V_{\pi}(s)$, which measures the relative importance of each action. By combining both streams, the estimated q-values are regularized across states of distinct value.

\vspace{0.5em}

\noindent\textbf{Noisy Networks:} Unlike the $\varepsilon$-greedy exploration strategy used in DQN, Noisy Networks \citep{Fortunato2018} introduce trainable noise into the network weights, allowing the model to learn useful perturbations for more efficient exploration. A noisy linear layer is defined as:
\begin{equation}
	y = (\mu^w + \sigma^w \odot \varepsilon^w)x + (\mu^b + \sigma^b \odot \varepsilon^b),
\end{equation}
where $\varepsilon^w, \varepsilon^b \sim \mathcal{N}(0, 1)$ are random noise variables, and $\mu^w, \mu^b, \sigma^w, \sigma^b$ are learnable parameters. Unlike simple random action selection, this method enables the agent to explore more intelligently by learning when and where uncertainty-driven exploration is beneficial. As a result, the agent is more likely to discover rewarding action sequences, even when immediate feedback is not possible.

\vspace{0.5em}

\noindent\textbf{Distributional RL:} Instead of learning the expected q-value, distributional RL \citep{Bellemare2017} learns to predict a full distribution of returns, leading to a richer understanding of uncertainty.

\vspace{0.5em}

\noindent\textbf{Multi-step learning:} This extension \citep{Sutton1998} considers rewards over multiple future timesteps rather than just the immediate next reward. Instead of updating the q-value based on a single-step reward, it accumulates rewards over multiple steps before making an update.

\subsection{Rationales behind value-based DRL for scheduling}
\label{sec:rationales}

As discussed before, value-based DRL offers several possible theoretical advantages over policy-gradient approaches in scheduling problems. Below, we elaborate on the key rationales for using value-based methods on scheduling problems, which ultimately motivated our investigation:
\begin{itemize}
	\item Off-policy learning and better sample efficiency: policy-gradient methods rely on on-policy learning, meaning they update their parameters using the most recent trajectories, which have no use afterwards. In contrast, value-based algorithms are off-policy. That is, they store past transitions in an experience replay buffer and reuse this data across multiple updates, providing better sample efficiency and learning more effectively from limited experiences. Particularly in real-world scheduling scenarios, keeping an experience replay buffer may be critical for efficiently using the gathered transitions, as collecting new experiences can be costly. 
	\item Stable training with target networks and experience replay: in DRL, training can become unstable due to correlated data and rapidly changing targets. Target networks mitigate this by using fixed target q-values that are updated periodically, preventing the algorithm from chasing a moving target. Furthermore, experience replays break the temporal correlation in sequential data by storing past transitions and sampling uniformly from this buffer. This is of particular interest in scheduling problems, where agents should estimate the best action on each step regardless of previous transitions. Together, these techniques improve stability and promote a more reliable convergence.
	\item Discrete action spaces: in scheduling problems, the number of actions is finite, resulting in discrete action spaces, where value-based methods are typically more efficient. This is because value-based algorithms can explicitly estimate and compare the q-values of all possible actions. In contrast, policy-gradient methods are known to be better suited for continuous action spaces, where enumerating all possible actions is not feasible. As a result, identifying the action with the highest q-value becomes non-trivial in these cases.
	\item Reduced variance in updates: in policy-gradient algorithms, the gradient of the expected return is estimated as:
 	\begin{equation}
		\nabla_{\theta} J(\theta) = \mathbb{E}_{\tau \sim \pi_{\theta}} \left[\sum_{t=0}^T \nabla_{\theta} \log \pi_{\theta}(a_t|s_t) G_t \right],
	\end{equation}
	which involves sampling multiple entire trajectories. When minimizing makespan, the return $G_t$ likely varies with the scheduling horizon, which can present wide ranges across training instances. This trajectory fluctuation, coupled with the stochasticity of the policy, can lead to high variance in gradient estimates. Value-based methods, on the other hand, learn via bootstrapping, meaning that value estimates are based on other value estimates, not full returns. This leads to biased but lower variance updates.
	\item Credit assignment and sparsity of rewards: scheduling problems, like other CO problems, involve actions with delayed and long-term consequences. For example, dispatching the wrong operation early in the schedule may lead to negative consequences that only become apparent much later. In addition, scheduling problems are typically featured by sparse rewards. At each decision point, the agent receives a reward of zero unless the partial makespan increases. Together, delayed consequences and sparse feedback lead to a credit assignment challenge, i.e., determining which actions are responsible for which outcomes. In this regard, many value-based extensions are designed to mitigate these challenges. PER, for example, focuses on sampling transitions with a higher temporal-difference error more frequently, which helps to propagate reward signals more effectively. Multi-step learning directly addresses the credit assignment challenge by carrying information about future rewards further back. Noisy networks are also helpful in this sense, as they inject trainable noise into the deep neural network parameters, enabling a more intelligent and adaptive exploration. 
\end{itemize}

\section{Model architecture}
\label{sec:architecture}

In our work, we follow the same model architecture designed by \citet{Song2023}, a heterogeneous GNN. Our choice for using this model as a baseline is two-fold: first, it is a well-consolidated model in the literature, being used as a benchmark across several works \citep{Wang2024,Yuan2024,Echeverria2025}; second, its flexibility allows it to be easily adapted to solve JSSP instances without any loss of generality. In this section, we describe all the details of the architecture used in this paper.

\subsection{MDP formulation}
\label{sec:mdp}

As discussed in Section~\ref{sec:rl}, RL approaches to CO are typically framed as an MDP. In this subsection, we present the specific MDP formulation for the JSSP and FJSP.

\vspace{0.5em}

\noindent\textbf{State:} At timestep $t$, the state $s_t$ is represented as a partial solution schedule $S(t)$, represented as a graph encoding the current status of all scheduled and unscheduled operations together with the machines and their processing status.

\vspace{0.5em}

\noindent\textbf{Action:} The action at timestep $t$ consists of selecting a valid operation-machine pair $(O_{ij}, M_k)$, where $O_{ij}$ is an assignable operation and $M_k \in \mathcal{M}_{ij}$ is one of its eligible machines. For selecting $(O_{ij}, M_k)$, the predecessor operations of $O_{ij}$ must all be completed, and each $M_k \in \mathcal{M}_{ij}$ must be idle at timestep $t$.In the JSSP, as there is no machine flexibility, the action is reduced to simply selecting the next operation to schedule. At each timestep, a mask is applied to restrict the agent to feasible actions. Specifically, machines that are still processing other operations, operations whose predecessors have not yet been completed, and operations that have already been processed are all excluded.

\vspace{0.5em}

\noindent\textbf{State transition:} After selecting an operation-machine pair, the operation is dispatched and added to the partial solution schedule. The machine availability is updated to reflect the execution of $O_{ij}$, and the operation is removed from the set of processable operations to prevent it from being selected again.

\vspace{0.5em}

\noindent\textbf{Reward:} Following the state transition, the agent receives a reward $r_t$, which is a scalar feedback signal reflecting the impact of the chosen action on the overall objective function. Specifically, the reward is defined as the negative increase in the makespan of the partial schedule caused by the selected assignment.

\subsection{Graph representation}
\label{sec:graph_representation}

Following many existing works \citep{Park2021, Song2023, Ho2024}, we represent the two problems described in Section~\ref{sec:prob_description} as a graph $\mathcal{G} = (\mathcal{V}, \mathcal{E})$. Here, $\mathcal{V}$ denotes the set of nodes and $\mathcal{E}$ the set of edges in an instance. The node set $\mathcal{V}$ consists of two subsets of different types of nodes: $\mathcal{O}$ is the set of operation nodes, which contains all operations from all jobs, along with two dummy nodes \textit{Start} and \textit{End}, which represent the beginning and end of the scheduling process; $\mathcal{M}$ is the set of machine nodes, containing all machines in the problem. The edge set $\mathcal{E}$ captures the relationships and constraints between nodes in the graph and can be divided into two types based on the nature of the connections. Operation-to-operation edges are conjunctive, directed arcs representing the processing sequence of each job $J_i \in \mathcal{J}$, forming paths from \textit{Start} to \textit{End} through the corresponding operations. In contrast, operation-machine edges are disjunctive, undirected arcs that connect each operation $O_{ij}$ to its set of candidate eligible machines $\mathcal{M}_{ij}$. Each of these edges is characterized by a feature representing the processing time $p_{ijk}$, which specifies the time required for operation $O_{ij}$ to be processed on machine $M_k$. In the JSSP, where machine assignments are fixed, each operation is connected by a single edge to its designated machine, reflecting the absence of flexibility in machine selection.

At each timestep $t$, the state $s_t$ is a partial solution encoded using the graph representation described above. The edge connections are dynamically updated to reflect changes in the production environment resulting from each machine-operation action selection. For example, if an operation $O_{ij}$ has multiple eligible machines $\mathcal{M}_{ij} = \{M_1, M_2, M_3\}$ and is dispatched to machine $M_2$, then all disjunctive edges connecting $O_{ij}$ to the other candidate machines $(M_1, M_3)$ are removed. To conform to the MDP formulation, each node in a partial solution must include features that encode all necessary state information, ensuring that the current solution schedule can be reconstructed independently of any preceding or future states. We detail each state feature for operation and machine nodes in \ref{sec:appA}.

\subsection{Graph neural network}
\label{sec:gnn}

\begin{figure*}[h]
	\centering
	\includegraphics[width=\textwidth]{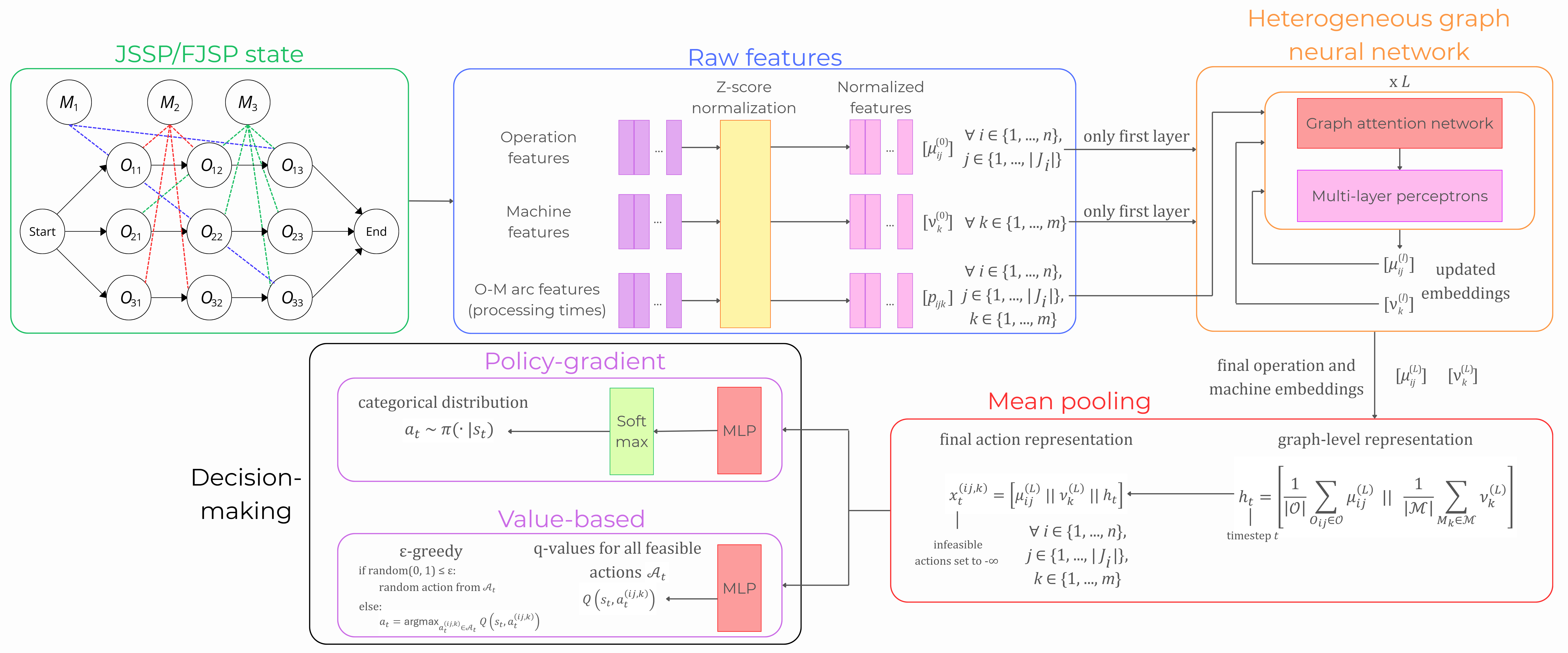}
	\caption{Unraveling the Rainbow architecture.}
	\label{fig_architecture}
\end{figure*}

GNNs are a powerful type of deep neural network specifically designed to learn from graph-structured data. Since many CO problems can be generally formulated as graphs, GNNs are well-suited to address them. Due to their ability to effectively capture complex dependencies and relational information, GNNs have become increasingly popular in the CO literature. In this work, we adapt the architecture used by \citet{Song2023}, a heterogeneous GNN. Each layer of the heterogeneous GNN projects every machine and operation node into a $d$-dimensional embedding. Machine and operation nodes are embedded using different mechanisms to account for their distinct topological roles within the scheduling graph: machine nodes are only connected to operations (via disjunctive arcs), while operation nodes are connected both to machines and to other operations (via conjunctive arcs representing precedence constraints). Below, we describe how each layer $l \in {1,...,L}$ of the GNN encodes machine and operation nodes. An overview of the model architecture can be seen in Figure~\ref{fig_architecture}.

\noindent\textbf{Machine nodes:} To compute the embedding of a machine node $M_k$, the model aggregates messages from its neighboring operations $\mathcal{N}_{t}(M_k)$, at timestep $t$. Each neighboring operation $O_{ij} \in \mathcal{N}_t(M_k)$ is associated with an operation-machine edge feature, namely the processing time $p_{ijk}$, which is incorporated into the calculation of attention coefficients. For each operation-machine pair, the model first linearly projects the operation, machine, and edge raw features into hidden spaces using learnable matrices $W_O \in \mathbb{R}^{d \times 6}$, $W_M \in \mathbb{R}^{d \times 3}$ and $W_E \in \mathbb{R}^{d \times 1}$, respectively. The attention score between operation $O_{ij}$ and machine $M_k$ is then computed using elementwise multiplication with learned head-specific attention vectors $a_O,a_M,a_E$, followed by a \textit{LeakyReLU} activation (with a negative slope = 0.2):
\begin{equation}
	e_{ijk} = \text{LeakyReLU} ((W_{O}^{(l)} \mu_{ij}^{(l-1)} \odot a_O) + (W_{M}^{(l)} \nu_{k}^{(l-1)} \odot a_M) + (W_{E}^{(l)} p_{ijk} \odot a_E)),
\end{equation}
where $\mu_{ij}^{(l-1)}$ and $\nu_k^{(l-1)}$ are the operation and machine representations from the previous layer. A distinction is made between the first and subsequent layers. In the first layer ($l=1$), $\mu_{ij}^{(0)} \in \mathbb{R}^6$ and $\nu_k^{(0)} \in \mathbb{R}^3$ correspond to the raw operation and machine features. These are projected into hidden spaces via $W_O^{(1)}$ and $W_M^{(1)}$. In subsequent layers ($l>1$), both $\mu_{ij}^{(l-1)}$ and $\nu_k^{(l-1)}$ are already $d$-dimensional embeddings, and so the model applies shared linear projections $W_O,W_M \in \mathbb{R}^{d \times d}$. We also note that, before the first GNN layer, all operation, machine and edge features are normalized using Z-score normalization.

A self-attention term is also computed for each machine node to capture its own context:
\begin{equation}
	e_{kk} = \text{LeakyReLU}((W_{M}^{(l)} \nu_{k}^{(l-1)} \odot a_M) + (W_{M}^{(l)} \nu_{k}^{(l-1)} \odot a_M)).
\end{equation}

All attention scores $\{e_{ijk}, e_{kk}\}$ are normalized by a softmax function to obtain attention weights $\alpha_{ijk}$ and $\alpha_{kk}$. Finally, the resulting embedding for machine $M_k$ is updated by a weighted combination of operation messages and its own self-message:
\begin{equation}
	\nu_{k}^{(l)} = \sigma(\alpha_{kk}W_{M}^{(l)} \nu_k^{(l-1)} + \sum_{O_{ij} \in \mathcal{N}_t(M_k)} \alpha_{ijk} (W_{O}^{(l)} \mu_{ij}^{(l-1)} + W_{E}^{(l)} p_{ijk})),
\end{equation}
where $\sigma(\cdot)$ is a sigmoid activation function.

\vspace{0.5em}

\noindent\textbf{Operation nodes:} The embedding of each operation node $O_{ij}$ is computed by aggregating information from four distinct sources: its predecessor $O_{i(j-1)}$, its successor $O_{i(j+1)}$, the set of neighboring machines $\mathcal{N}_t (O_{ij})$, and $O_{ij}$ itself. First, the embeddings of all machines in $\mathcal{N}_t(O_{ij})$, (that is, all machines capable of processing $O_{ij}$ are aggregated using an element-wise sum:
\begin{equation}
	\bar{\nu}_{ij}^{(l)} = \sum_{M_k \in \mathcal{N}_t (O_{ij})} \nu_k^{(l)}.
\end{equation}

Each of the four inputs --- $\mu_{i(j-1)}^{(l-1)}, \ \mu_{i(j+1)}^{(l-1)}, \ \bar{\nu}_{ij}^{(l)} \ \text{and} \ \mu_{ij}^{(l-1)}$ --- is processed by a dedicated MLP. Each MLP consists of two hidden layers with ELU activations and independently learns relation-specific transformations for the respective input type. The resulting representations are concatenated and passed through a final projection MLP:
\begin{equation}
	\mu_{ij}^{(l)} = \text{MLP}_{\theta_0}(
	\text{ELU} [
	\text{MLP}_{\theta_1}(\mu_{i(j-1)}^{(l-1)}) \, || \, 
	\text{MLP}_{\theta_2}(\mu_{i(j+1)}^{(l-1)}) \, || \, 
	\text{MLP}_{\theta_3}(\bar{\nu}_{ij}^{(l)}) \, || \, 
	\text{MLP}_{\theta_4}(\mu_{ij}^{(l-1)})
	]
	)
\end{equation}

\vspace{0.5em}

\noindent\textbf{Mean pooling:} After the final GNN layer, the embeddings of all machine and operation nodes are aggregated separately using mean pooling. The resulting pooled vectors, one for machines and one for operations, each in $\mathbb{R}^d$, are then concatenated to form a single graph-level representation:
\begin{equation}
	h_t = [\frac{1}{|\mathcal{O}|}\sum_{O_{ij}\in\mathcal{O}}\mu_{ij}^{(L)} \ || \ \frac{1}{|\mathcal{M}|}\sum_{M_k \in \mathcal{M}} \nu_k^{(L)}] \in \mathbb{R}^{2d}.
\end{equation}

The graph-level representation $h_t$, together with the final node embeddings produced by the GNN, is then used to compute either action probabilities (in policy-gradient algorithms) or q-values (in value-based methods). In both cases, an action corresponds to an operation-machine pair. At each timestep $t$, each feasible action $a_t^{(ij,k)} = (O_{ij}, M_k) \in \mathcal{A}_t$ is evaluated using the corresponding operation, machine and graph-level embeddings:
\begin{equation}
	x_t^{(ij, k)} = [\mu_{ij}^{(L)} \ || \ \nu_k^{(L)} \ || h_t] \in \mathbb{R}^{4d}.
\end{equation}

\vspace{0.5em}

\noindent\textbf{Policy-gradient methods:} In the policy-gradient setting, the probability of selecting action $a_t^{(ij,k)}$ given state $s_t$ is computed through an MLP, followed by a softmax over all feasible actions:
\begin{equation}
	\pi (a_t^{(ij, k)} \ | \ s_t) = \frac{\text{exp} \ (\text{MLP}_{\theta_5}(x_t^{(ij, k)}))}{\displaystyle \sum_{a_t^{(i'j', k')} \in \mathcal{A}_t} \exp(\text{MLP}_{\theta_5}(x_t^{(i'j',k')}))}.
\end{equation}

This MLP consists of two linear layers with tanh activations, and outputs a scalar score (logit) for each action. The softmax function then converts these logits into a normalized probability distribution over feasible actions. During training, the agent samples actions from the categorical distribution defined by $\pi(\cdot|s_t)$, enabling stochastic exploration over feasible actions:
\begin{equation}
	a_t \sim \pi(\cdot | s_t).
\end{equation}

\vspace{0.5em}

\noindent\textbf{Value-based methods:} In value-based DRL, the q-value associated with each action is estimated as:
\begin{equation}
	Q(s_t, a_t^{(ij, k)}) = \text{MLP}_{\theta_5} (x_t^{(ij, k)}),
\end{equation}
where an MLP with the same two-layer architecture is employed. During training, action selection follows an $\varepsilon$-greedy strategy: with probability $\varepsilon$, a random feasible action is chosen uniformly from $\mathcal{A}_t$, otherwise, the action with the highest q-value is selected as:
\begin{equation}
	a_t = \text{argmax}_{a_t^{(ij, k)} \in \mathcal{A}_t} \ Q(s_t, a_t^{(ij, k)}).
\end{equation}

Note that in both policy-gradient and value-based algorithms, a masking mechanism is applied ensuring that infeasible actions are not selected. In policy-gradient methods, the logit of an infeasible action $a_t^{(ij,k)}$ is set to $-\infty$ before the softmax is applied, yielding a probability of zero. In value-based methods, the Q-value $(s_t,a_t^{(ij,k)})$ of any infeasible action is similarly set to $-\infty$ before applying the argmax, ensuring it is never chosen.

\section{Experimental results}
\label{sec:experiments}

\subsection{Experimental settings and baselines}
\label{sec:exp_settings}

Following prior work \citep{Hessel2017RainbowCI,Obando2020}, we evaluate the impact of adding each algorithmic extension to the DQN. For both the JSSP and FJSP, each algorithm was trained separately on multiple instance sizes (6x6, 10x5, 20x5, 15x10, 20x10), resulting in one trained model per algorithm-instance size combination. Value-based methods were trained for 3000 episodes on the smaller instances (6x6, 10x5, 20x5), and 5000 episodes on the larger ones (15x10, 20x10). Policy-gradient algorithms were trained for the same number of episodes. To mitigate overfitting, new training instances were generated at the start of every episode. Details of the instance generation process are provided in the following subsection. Throughout training, we periodically performed a validation step. For value-based algorithms, this occurred whenever the target network was updated. In each validation step, the current online network (or policy, for policy-gradient algorithms) was evaluated on a set of 100 instances, generated using the same procedure as for training.

To conduct all testing experiments, we selected the model that achieved the lowest average makespan across all validation steps during training. We then evaluated each different model on 100 distinct testing instances per instance size. In addition to the randomly generated instances, we also evaluated the models on standard public benchmark datasets. For the JSSP, we used the datasets from Taillard, Demirkol, and Lawrence \citep{Taillard1993,Demirkol1998,Lawrence1984}. For the FJSP, we adopted the benchmarks from Brandimarte, Dauzère and Paulli, Chambers and Barnes, and Hurink \citep{Brandimarte1993,Dauzere1997,Chambers1996,Hurink1994}. In all experiments, inference was performed using a greedy decoding, where at each timestep the action with the highest probability (or Q-value) is selected. 

To benchmark the results of all algorithms, we used Google OR-Tools \citep{Furnon2024}, a powerful constraint programming algorithm widely used in the literature. We set a time limit of 45 minutes for solving each individual testing instance. For public benchmark problems, we used the best-known solutions in the literature \citep{Behnke2012,Zhang2020,Yuan2024}.

All models were trained on an Ubuntu v24.04 machine with 32GB of RAM, an Intel Core i9-13900 and an NVIDIA GeForce RTX 4060. All hyperparameters are detailed in \ref{sec:appB}.

\subsection{Instance generation}
\label{sec:instance_gen}

To generate all training, validation and testing instances, we followed different procedures for the JSSP and FJSP due to the distinct nature of each problem. For the JSSP, we used the same method as \citet{Zhang2020}. The procedure is straightforward since the JSSP does not involve machine flexibility. Each job consists of exactly $m$ operations, where $m$ is the number of machines in the problem. For every job, operations are assigned to a randomly selected permutation of the available machines, meaning each operation is assigned to a different machine. The processing times for these operations are then drawn uniformly from the range [1, 99].

For the FJSP, we followed the approach used by \citet{Song2023}. Each job $J_i$ has a number of operations $N_i$. The value of $N_i$ is uniformly sampled on different ranges based on the number of machines in the problem: [4, 6] for instances with 5 machines, [5, 7] for 6 machines and [8, 12] for 10 machines. For each operation $O_{ij}$, a set of candidate machines $M_{ij}$ is sampled, with the number of candidates ranging from 1 to $m$ . An average processing time $\bar{p}_{ij}$ for each operation is also drawn uniformly from the interval [1, 20]. The actual processing time for operation $O_{ij}$ on each candidate machine $M_k$ is then sampled from a uniform distribution over the range [0.8$\bar{p}_{ij}$,1.2$\bar{p}_{ij}$].

\subsection{Results on randomly generated instances}
\label{sec:random_results}

In this subsection, we analyze our results through the lens of questions 1, 2, and 3 outlined at the end of Section~\ref{sec:intro}: 1) \textit{Can value-based DRL algorithms compete with policy-gradient ones in job-shop scheduling problems?} 2) \textit{Do value-based algorithms, like Rainbow and its individual components, offer significant improvements over the DQN algorithm when solving job-shop scheduling problems?} 3) \textit{How does the performance of value-based methods vary when applied to problems with higher routing flexibility, such as the FJSP, compared to those with a fixed routing structure, like the JSSP?}

\begin{figure*}[h]
	\centering
	\includegraphics[width=\textwidth]{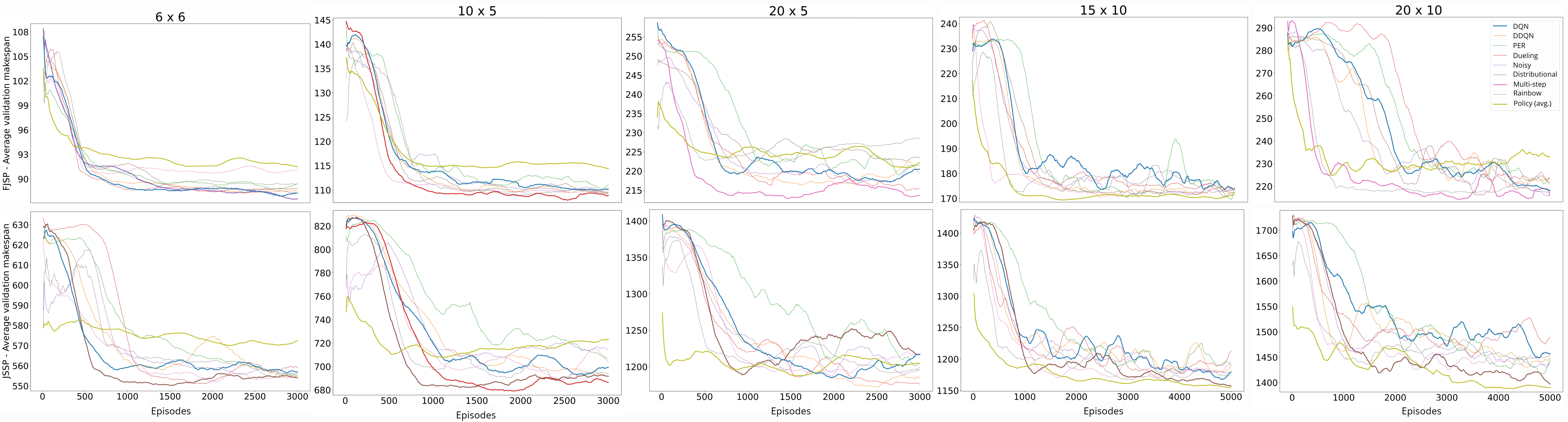}
	\caption{Average validation makespan throughout training for each value-based algorithm, shown separately for all instance sizes. Each subplot corresponds to a specific instance size. The top row displays results for the FJSP, and the bottom row for the JSSP.}
	\label{fig_val_value}
\end{figure*}

\begin{figure*}[h]
	\centering
	\includegraphics[width=\textwidth]{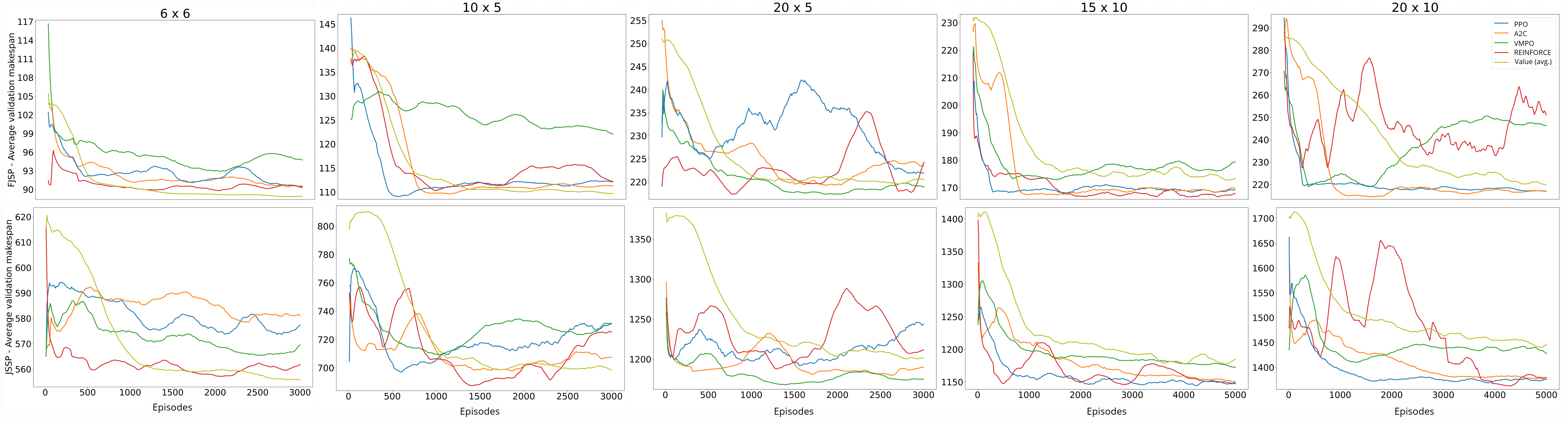}
	\caption{Average validation makespan throughout training for each policy-gradient algorithm, shown separately for all instance sizes. Each subplot corresponds to a specific instance size. The top row displays results for the FJSP, and the bottom row for the JSSP.}
	\label{fig_val_policy}
\end{figure*}

To answer the first question, we began by comparing the training convergence of models trained with all the different value-based and policy-gradient algorithms outlined in Section~\ref{sec:prelim} (shown in Figures~\ref{fig_val_value} and \ref{fig_val_policy}). Each curve shows the validation score (average makespan across the validation dataset), which was evaluated every 10 episodes. To facilitate visualization, Figure~\ref{fig_val_value} displays each value-based method, and the average across policy-gradient algorithms, while Figure~\ref{fig_val_policy} shows the exact opposite (individual policy-gradient performances with the average of all value-based). We find that, in smaller instances, on both the JSSP and FJSP, most value-based algorithms outperformed policy-gradient methods, which struggled to converge effectively. For example, on 6x6 JSSP instances, all value-based methods were able to converge effectively around an average makespan of 550-560, while policy-gradient stayed between 570-580. However, this pattern shifted in larger problems (15x10 and 20x10), where policy methods demonstrated a faster convergence during the first 1500 episodes.

\begin{figure*}[h]
	\centering
	\includegraphics[width=\textwidth]{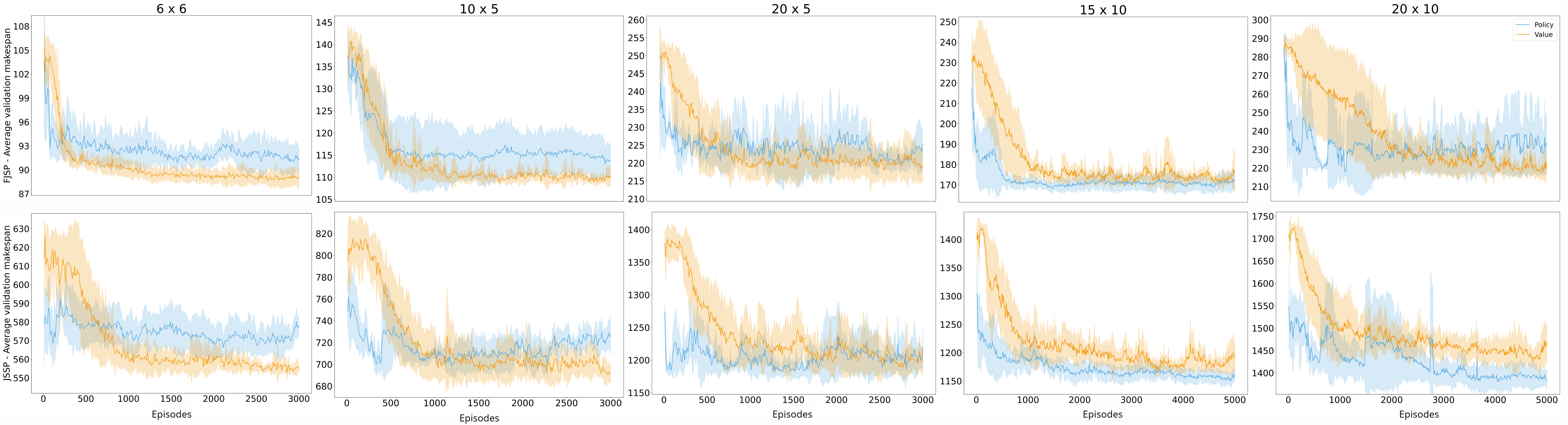}
	\caption{Average validation makespan throughout training for each class of DRL algorithms, shown separately for all instance sizes. Each subplot corresponds to a specific instance size. The top row displays results for the FJSP, and the bottom row for the JSSP. Shaded regions represent one standard deviation above and below the mean.}
	\label{fig_shaded_val}
\end{figure*}

To confirm the observed trends, Figure~\ref{fig_shaded_val} presents the evolution of the validation score for each class of DRL methods, with shaded regions indicating one standard deviation above and below the mean. Here, we can clearly notice the higher variance associated with policy-gradient methods, which often demonstrated a much more erratic behavior compared to value-based algorithms. In contrast, value-based methods showed more stable convergence and lower standard deviations. Despite their instability, across all instance sizes, policy-gradient methods tended to converge faster within the first 500 episodes (even on smaller instances, where value-based prevailed). A plausible explanation is that value-based methods are more prone to bias. These methods learn via bootstrapping, which means they update their predictions using their own previous estimates rather than the actual observed outcomes. In other words, the network learns from approximations of future rewards instead of the true returns.

Concretely, during each minibatch update, the parameters of the online network are adjusted by minimizing the mean squared error between its predicted q-values and the target q-values, provided by the separate target network. Since both sides of this update depend on learned (and therefore imperfect) estimates, this process introduces bias. As a result, early in training, when the action-value function is still poorly initialized, these updates may be inaccurate, and it can take longer for the estimates to become reliable, which can slow down learning in the initial phase. This suggests that, under time-constrained training budgets, policy-gradient methods may offer faster performance gains. However, for extended training sessions, the greater stability and lower variance of value-based algorithms might make them a more favorable choice for scheduling problems.

Then, to assess if the trends observed during training held during inference, we evaluated all models on 100 randomly generated testing instances of the same size as the training instances. We also report results for 30x10 and 40x10 instances using models trained on problems with 20 jobs and 10 machines, similar to \citet{Song2023}. Tables~\ref{table_results_jssp} and \ref{table_results_fjsp} report the average makespan across all testing instances for both the JSSP and FJSP, respectively, as well as the performance gap relative to OR-Tools. For both value-based and policy-gradient methods, the best-performing results are highlighted in \textbf{bold}, along with an indication of which category of DRL algorithms performed better. 

Overall, we observed that results varied between problem types (JSSP and FJSP) and instance sizes. On the JSSP, most value-based methods performed better on smaller and medium sized instances (up to 10x5). However, as the problem size increased, policy-gradient methods showed a better performance than value-based. As a result, policy-gradient algorithms achieved a lower average performance gap overall on the JSSP. In contrast, on the FJSP, value-based methods consistently outperformed policy-gradient approaches across all instance sizes, resulting in superior average performance.

\begin{table*}[h]
	\caption{Results on datasets of 100 randomly generated JSSP instances.}
	\label{table_results_jssp}
	\centering
	\resizebox{\textwidth}{!}{
		\renewcommand{\arraystretch}{1.2}
		\begin{tabular}{c|cc|cc|cc|cc|cc|cc|cc|cc}
			\hline
			Algorithm & \multicolumn{2}{c|}{6x6} 
			& \multicolumn{2}{c|}{10x5} 
			& \multicolumn{2}{c|}{20x5} 
			& \multicolumn{2}{c|}{15x10} 
			& \multicolumn{2}{c|}{20x10} 
			& \multicolumn{2}{c|}{30x10} 
			& \multicolumn{2}{c|}{40x10} 
			& \multicolumn{2}{c}{Average} \\
			& $C_{\max}$ & Gap 
			& $C_{\max}$ & Gap 
			& $C_{\max}$ & Gap 
			& $C_{\max}$ & Gap 
			& $C_{\max}$ & Gap 
			& $C_{\max}$ & Gap 
			& $C_{\max}$ & Gap 
			& $C_{\max}$ & Gap \\
			\hline
			OR-Tools & 499.72 & -- & 622.45 & -- & 1153.08 & -- & 1001.53 & -- & 1215.81 & -- & 1751.77 & -- & 2266.54 & -- & 1215.84 & -- \\
			\hline
			PPO & 567.22 & 13.51\% & 685.63 & 10.15\% & 1179.40 & 2.28\% & \textbf{1157.72} & \textbf{15.60\%} & 1379.35 & 13.45\% & 1871.66 & 6.84\% & 2369.48 & 4.54\% & 1315.78 & 9.48\% \\
			A2C & 568.21 & 13.71\% & 687.48 & 10.45\% & \textbf{1174.74} & \textbf{1.88\%} & 1170.80 & 16.90\% & 1384.80 & 13.90\% & 1885.14 & 7.61\% & 2371.73 & 4.64\% & 1320.41 & 9.87\% \\
			REINFORCE & \textbf{554.48} & \textbf{10.96\%} & \textbf{675.98} & \textbf{8.60\%} & 1185.76 & 2.83\% & 1160.94 & 15.92\% & \textbf{1372.47} & \textbf{12.89\%} & \textbf{1864.50} & \textbf{6.44\%} & \textbf{2332.32} & \textbf{2.90\%} & \textbf{1306.64} & \textbf{8.65\%} \\
			V-MPO & 567.18 & 13.50\% & 705.11 & 13.28\% & 1175.76 & 1.97\% & 1196.90 & 19.51\% & 1414.14 & 16.31\% & 1915.05 & 9.32\% & 2385.62 & 5.25\% & 1337.11 & 11.31\% \\
			\hline
			Policy (avg.) & 564.27 & 12.92\% & 688.55 & 10.62\% & \textbf{1178.91} & \textbf{2.24\%} & \textbf{1171.59} & \textbf{16.98\%} & \textbf{1387.69} & \textbf{14.14\%} & \textbf{1884.09} & \textbf{7.55\%} & \textbf{2364.79} & \textbf{4.33\%} & \textbf{1319.98} & \textbf{9.83\%} \\
			\hline
			DQN & 556.11 & 11.28\% & 679.14 & 9.11\% & 1174.39 & 1.85\% & 1172.11 & 17.03\% & 1434.35 & 17.97\% & 1942.55 & 10.89\% & 2393.91 & 5.62\% & 1336.08 & 10.54\% \\
			DDQN & \textbf{550.09} & \textbf{10.08\%} & 677.58 & 8.86\% & \textbf{1173.28} & \textbf{1.75\%} & 1184.78 & 18.30\% & 1406.48 & 15.68\% & 1932.04 & 10.29\% & 2397.14 & 5.76\% & 1331.63 & 10.10\% \\
			PER & 554.24 & 10.91\% & 692.54 & 11.26\% & 1183.00 & 2.59\% & 1176.33 & 17.45\% & 1407.38 & 15.76\% & 1918.13 & 9.50\% & 2402.53 & 6.00\% & 1333.45 & 10.50\% \\
			Dueling & 550.98 & 10.26\% & \textbf{671.79} & \textbf{7.93\%} & 1177.63 & 2.13\% & \textbf{1168.10} & \textbf{16.63\%} & 1428.72 & 17.51\% & 1948.23 & 11.21\% & 2427.09 & 7.08\% & 1338.93 & 10.39\% \\
			Noisy & 553.86 & 10.83\% & 685.80 & 10.18\% & 1196.48 & 3.76\% & 1179.84 & 17.80\% & 1403.22 & 15.41\% & 1899.48 & 8.43\% & 2414.60 & 6.53\% & 1333.33 & 10.42\% \\
			Distributional & 551.75 & 10.41\% & 673.98 & 8.28\% & 1185.64 & 2.82\% & 1172.27 & 17.05\% & \textbf{1382.60} & \textbf{13.72\%} & \textbf{1893.85} & \textbf{8.11\%} & \textbf{2361.03} & \textbf{4.17\%} & \textbf{1317.30} & \textbf{9.22\%} \\
			Multi-step & 555.71 & 11.20\% & 681.32 & 9.46\% & 1181.00 & 2.42\% & 1184.05 & 18.22\% & 1415.95 & 16.46\% & 1915.18 & 9.33\% & 2374.60 & 4.77\% & 1329.69 & 10.27\% \\
			Rainbow & 557.16 & 11.49\% & 682.43 & 9.64\% & 1195.68 & 3.69\% & 1185.33 & 18.35\% & 1407.64 & 15.78\% & 1927.16 & 10.01\% & 2396.33 & 5.73\% & 1335.96 & 10.67\% \\
			\hline
			Value (avg.) & \textbf{553.74} & \textbf{10.81\%} & \textbf{680.57} & \textbf{9.34\%} & 1183.39 & 2.63\% & 1177.85 & 17.60\% & 1410.79 & 16.04\% & 1922.08 & 9.72\% & 2395.90 & 5.71\% & 1332.05 & 10.26\% \\
			\hline
		\end{tabular}
	}
\end{table*}

\begin{table*}[h]
	\caption{Results on datasets of 100 randomly generated FJSP instances.}
	\label{table_results_fjsp}
	\centering
		\resizebox{\textwidth}{!}{
		\renewcommand{\arraystretch}{1.2}
		\begin{tabular}{c|cc|cc|cc|cc|cc|cc|cc|cc}
			\hline
			Algorithm & \multicolumn{2}{c|}{6x6} 
			& \multicolumn{2}{c|}{10x5} 
			& \multicolumn{2}{c|}{20x5} 
			& \multicolumn{2}{c|}{15x10} 
			& \multicolumn{2}{c|}{20x10} 
			& \multicolumn{2}{c|}{30x10} 
			& \multicolumn{2}{c|}{40x10} 
			& \multicolumn{2}{c}{Average} \\
			& $C_{\max}$ & Gap 
			& $C_{\max}$ & Gap 
			& $C_{\max}$ & Gap 
			& $C_{\max}$ & Gap 
			& $C_{\max}$ & Gap 
			& $C_{\max}$ & Gap 
			& $C_{\max}$ & Gap 
			& $C_{\max}$ & Gap \\
			\hline
			OR-Tools & 72.70 & -- & 96.80 & -- & 191.80 & -- & 153.22 & -- & 206.99 & -- & 307.11 & -- & 405.75 & -- & 204.91 & -- \\
			\hline
			PPO & \textbf{85.16} & \textbf{17.14\%} & \textbf{110.69} & \textbf{14.35\%} & 213.38 & 11.25\% & 167.65 & 9.42\% & 217.60 & 5.13\% & 315.56 & 2.75\% & 418.43 & 3.13\% & 218.35 & 9.02\% \\
			A2C & 85.59 & 17.73\% & 111.82 & 15.52\% & 214.51 & 11.84\% & \textbf{166.32} & \textbf{8.55\%} & \textbf{215.76} & \textbf{4.24\%} & \textbf{312.74} & \textbf{1.83\%} & \textbf{416.70} & \textbf{2.70\%} & \textbf{217.63} & \textbf{8.92\%} \\
			REINFORCE & 85.44 & 17.52\% & 112.61 & 16.33\% & \textbf{212.28} & \textbf{10.68\%} & 167.25 & 9.16\% & 217.10 & 4.88\% & 314.85 & 2.59\% & 417.81 & 2.97\% & 218.19 & 9.16\% \\
			V-MPO & 86.72 & 19.28\% & 123.17 & 27.24\% & 213.60 & 11.37\% & 172.30 & 12.45\% & 218.08 & 5.36\% & 315.25 & 2.65\% & 419.17 & 3.31\% & 221.18 & 11.67\% \\
			\hline
			Policy (avg.) & 85.73 & 17.92\% & 114.57 & 18.36\% & 213.44 & 11.28\% & \textbf{168.38} & \textbf{9.89\%} & 217.13 & 4.90\% & 314.60 & 2.45\% & 418.03 & 3.03\% & 218.84 & 9.69\% \\
			\hline
			DQN & 83.39 & 14.70\% & 112.21 & 15.92\% & 213.26 & 11.19\% & 168.36 & 9.88\% & 215.95 & 4.33\% & 314.25 & 2.32\% & 417.77 & 2.96\% & 217.88 & 8.76\% \\
			DDQN & 82.51 & 13.49\% & 110.64 & 14.30\% & 211.14 & 10.08\% & 169.80 & 10.82\% & 217.79 & 5.22\% & 314.21 & 2.31\% & 417.29 & 2.84\% & 217.63 & 8.44\% \\
			PER & 83.70 & 15.13\% & 110.40 & 14.05\% & 213.45 & 11.29\% & 168.16 & 9.75\% & 217.20 & 4.93\% & 313.32 & 2.02\% & 415.01 & 2.28\% & 217.32 & 8.49\% \\
			Dueling & \textbf{82.20} & \textbf{14.44\%} & \textbf{110.10} & \textbf{13.74\%} & 210.44 & 9.72\% & 170.79 & 11.47\% & 220.34 & 6.45\% & 315.86 & 2.85\% & 419.50 & 3.39\% & 218.46 & 8.87\% \\
			Noisy & 82.30 & 13.20\% & 110.87 & 14.54\% & 212.50 & 10.79\% & 168.51 & 9.98\% & 216.33 & 4.51\% & 312.88 & 1.88\% & 414.58 & 2.18\% & \textbf{216.85} & \textbf{8.15\%} \\
			Distributional & 82.91 & 14.04\% & 111.20 & 14.88\% & 217.07 & 13.18\% & \textbf{166.92} & \textbf{8.94\%} & \textbf{215.05} & \textbf{3.89\%} & 315.24 & 2.65\% & 426.45 & 5.10\% & 219.26 & 8.95\% \\
			Multi-step & 85.59 & 17.73\% & 110.87 & 14.54\% & \textbf{209.15} & \textbf{9.05\%} & 169.54 & 10.65\% & 215.61 & 4.16\% & \textbf{311.00} & \textbf{1.27\%} & \textbf{413.95} & \textbf{2.02\%} & 216.53 & 8.49\% \\
			Rainbow & 84.13 & 15.72\% & 112.10 & 15.81\% & 216.68 & 12.97\% & 170.52 & 11.29\% & 217.24 & 4.95\% & 314.56 & 2.43\% & 418.24 & 3.08\% & 219.07 & 9.46\% \\
			\hline
			Value (avg.) & \textbf{83.34} & \textbf{14.81\%} & \textbf{111.05} & \textbf{14.72\%} & \textbf{212.96} & \textbf{11.03\%} & 169.07 & 10.35\% & \textbf{216.94} & \textbf{4.80\%} & \textbf{313.91} & \textbf{2.22\%} & \textbf{417.85} & \textbf{2.98\%} & \textbf{217.87} & \textbf{8.70\%} \\
			\hline
		\end{tabular}
	}
\end{table*}

\begin{figure*}[h]
	\centering
	\includegraphics[width=\textwidth]{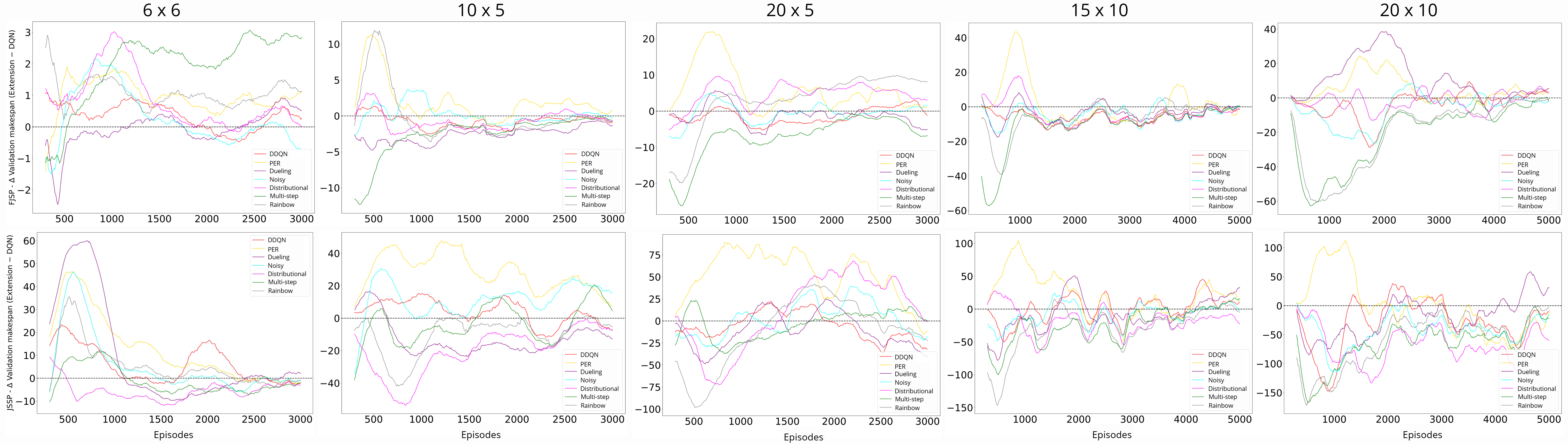}
	\caption{Difference in validation makespan over the course of training between each algorithmic component and DQN, shown separately for each instance size. Each subplot corresponds to a specific instance size. The top row displays results for the FJSP, and the bottom row for the JSSP.}
	\label{fig_dqn_vs_rainbow}
\end{figure*}

To answer the second question, we examined the training convergence of Rainbow and its individual components compared to DQN. Figure~\ref{fig_dqn_vs_rainbow} illustrates the difference in average validation makespan between each extension and the DQN, computed at each target network update. A positive difference indicates that DQN achieved a lower (and therefore better) makespan than the corresponding extension, whereas a negative value indicates that the extension outperformed DQN.

Overall, some algorithmic components showed improved convergence behavior than DQN, notably Distributional RL and Multi-step learning (see for example FJSP – 10x5, 15x10, 20x10, JSSP – 6x6, 15x10, 20x10). However, convergence speed varied considerably across components. For example, the PER initially converged slower than the DQN during the first 1000 to 2000 episodes, whereas the Multi-step learning model showed a faster early convergence. Another interesting aspect is that the full Rainbow implementation did not appear to generate any significant improvements over the best-performing individual components, indicating that, in fact, its application beyond the ALE scope is not straightforward.

Similarly to question 1, we also evaluated the inference performance of each separate value-based algorithm on the testing datasets (also reported in Tables~\ref{table_results_jssp} and \ref{table_results_jssp}).

On the JSSP, Distributional RL and Multi-step learning consistently outperformed DQN, especially as problem size increased, while other extensions did not show notable improvements. On the FJSP, the Multi-step learning and Noisy Networks were the most effective, outperforming the DQN in 5 out of the 7 tested instance sizes. For smaller problems (6x6 and 10x5), the Dueling Network led to lower average makespan than the rest, but showed a significant performance drop in larger instances.

Besides the credit assignment problem, which has been discussed previously in subsection~\ref{sec:rationales}, another possible reason for the success of the Multi-step learning extension stems from its ability to partially address the challenge of sparse rewards. In our scheduling environment, rewards are defined as the negative change in makespan. When a scheduling action does not immediately increase the current makespan, the associated reward is zero. Multi-step targets, by aggregating rewards over multiple future timesteps, help to better propagate the learning signals, making value estimation potentially more effective in this setting.

The performance of noisy networks was more pronounced on the FJSP likely due to its larger and more complex action space. In this regard, the stochastic exploration mechanism introduced by noisy networks may be more effective than the traditional $\varepsilon$-greedy strategy, allowing the agent to escape local optima.

Another interesting finding was the difference in performance for the Distributional RL extension, which was much better on the JSSP compared to the FJSP. Distributional RL may mitigate the harder credit assignment problem on the JSSP by modeling the full distribution of returns, not just the mean expected value. By capturing the range and variability of potential outcomes, it can provide the agent with richer feedback, allowing it to better account for the long-term uncertainty introduced by early decisions, as will be further discussed in the following paragraphs.

Overall, while some Rainbow extensions demonstrated promise, the cumulative impact of Rainbow extensions in scheduling problems appears less substantial than what has been observed in classic benchmarks like the ALE. In most cases, performance gains over DQN were incremental, raising concerns about whether the added complexity and implementation overhead of these algorithms are justified for CO problems. Nevertheless, despite their relatively modest gains in this domain, they remained competitive, and in many instances, even outperformed policy-gradient algorithms.

The third question addresses problem flexibility and its influence on algorithm performance. Results suggest a clear distinction between both problem variants. For the JSSP, with fixed operation routes and a more restricted action space, value-based methods underperformed compared to policy-gradient, especially as problem size increases. In contrast, results on the FJSP were much more promising. In this setting, most value-based approaches outperformed the policy-gradient ones. This difference may be attributed to the machine flexibility component inherent to the FJSP. 

Although the FJSP can be considered to have a larger state-action space than the JSSP due to the machine flexibility (which could, theoretically, make value-based algorithms less effective), the credit assignment challenge tends to be more severe in the JSSP. In the JSSP, operations must follow a fixed routing with no flexibility in machine choice, meaning that early mistakes in dispatching decisions can propagate through the rest of the schedule, making the credit assignment more challenging. This explains why the Multi-step learning and Distributional RL extensions performed generally better on the JSSP than other value-based algorithms. In contrast, the machine flexibility component of the FJSP allows the agent to recover from early suboptimal decisions, making the credit assignment easier since the impact of any single action is typically less catastrophic.

Likewise, we also observed from our experiments that the proportion of sparse rewards was a bit higher on JSSP instances than FJSP, especially at larger sizes. While some value-based extensions are designed to mitigate the problem of sparse rewards, value-based algorithms fundamentally rely on learning from individual transitions and on bootstrapping, updating the Q-value of a state-action pair based on the estimated value of the subsequent state. As many intermediate states produce a reward of zero, the signal for propagating the reward  can become weak. 

Overall, our findings indicate that value-based and policy-gradient methods can have different strengths depending on the nature of the problem, particularly in terms of structure, flexibility, and action space complexity.

\subsection{Cross-size and cross-distribution generalization results}
\label{sec:oos_results}

In this section, we address the fourth question presented in Section~\ref{sec:intro}: 4) \textit{How well do different value-based algorithms generalize across instances with varying sizes and distributions?} To investigate this, we conduct two analyses to evaluate the cross-size and cross-distribution generalization of all previously discussed models. Cross-size generalization refers to how well a model performs on problem instances that differ in size from those seen during training. Typically, this is measured by using much larger instances. Cross-distribution generalization, on the other hand, evaluates a model’s ability to handle instances generated from different distributions than those seen during training.

We begin by examining how well models trained on relatively small instances (6x6) generalize to substantially larger problems (20x10, 30x10 and 40x10). To extend this analysis, we also generated and solved 20 additional testing instances of sizes 50x20 and 100x20, for both the FJSP and JSSP. Like the previous instances, we ran OR-Tools for 45 minutes per instance. 

The motivation behind using models trained on small instances stems from the fact that their training times are much faster compared to models trained on larger instances. From our experiments, models trained on 6x6 instances required around 0.6 seconds per episode, whereas models trained on 20x10 instances required around 6 seconds. The cross-size generalization results are displayed in Table~\ref{table_cross_size}. Across all instance sizes, for both the JSSP and FJSP, value-based algorithms displayed a better generalization than policy-gradient methods. The improved generalization of value-based models suggests that the learned q-functions capture more transferable representations across larger problem scales.

\begin{table*}[h]
	\caption{Cross-size generalization results.}
	\label{table_cross_size}
	\centering
	\resizebox{\textwidth}{!}{
	\renewcommand{\arraystretch}{1.2}
		\begin{tabular}{cc|cc|cc|cc|cc|cc}
			\hline
			\multicolumn{2}{c|}{\multirow{2}{*}{Method}} & \multicolumn{2}{c|}{20$\times$10} & \multicolumn{2}{c|}{30$\times$10} & \multicolumn{2}{c|}{40$\times$10} & 
			\multicolumn{2}{c|}{50$\times$20} & 
			\multicolumn{2}{c}{100$\times$20} \\
			\multicolumn{2}{c|}{} & $C_{max}$ & Gap & $C_{max}$ & Gap & $C_{max}$ & Gap & $C_{max}$ & Gap & $C_{max}$ & Gap \\
			\hline
			\multicolumn{2}{c|}{OR-Tools} & 206.99 & - & 307.11 & - & 405.75 & - & 709.15 & - & 1330.40 & - \\
			\hline
			\multirow{14}{*}{FJSP} & PPO & 225.96 & 9.16\% & 319.40 & 4.00\% & 420.19 & 3.56\% & 549.90 & -22.46\% & 1041.55 & -21.71\% \\
			& A2C & 256.84 & 24.08\% & 350.30 & 14.06\% & 449.37 & 10.75\% & 607.60 & -14.32\% & 1093.35 & -17.82\% \\
			& REINFORCE & 250.60 & 21.07\% & 336.02 & 9.41\% & 444.30 & 9.50\% & 571.55 & -19.40\% & 1077.60 & -19.00\% \\
			& V-MPO & 224.95 & 8.68\% & 319.08 & 3.90\% & 421.68 & 3.93\% & 561.30 & -20.85\% & 1048.30 & -21.20\% \\
			\cline{2-12}
			& Policy (avg.) & 239.59 & 15.75\% & 331.20 & 7.84\% & 433.89 & 6.94\% & 572.59 & -19.26\% & 1065.20 & -19.93\% \\
			\cline{2-12}
			& DQN & 252.69 & 22.08\% & 347.25 & 13.07\% & 445.13 & 9.70\% & 608.65 & -14.17\%  & 1097.09 &-17.54\% \\
			& DDQN & 235.57 & 13.81\% & 342.38 & 11.48\% & 459.36 & 13.21\% & 566.90 & -20.06\% & 1035.20 & -22.19\% \\
			& PER & 221.91 & 7.21\% & 323.01 & 5.18\% & 427.37 & 5.33\% & 533.75 & -24.73\% & 1075.20 & -19.18\% \\
			& Dueling & 219.41 & 6.00\% & 316.11 & 2.93\% & 418.23 & 3.08\% & 541.75 & -23.61\% & 1031.15 & -22.49\% \\
			& Noisy & 225.11 & 8.75\% & 325.54 & 6.00\% & 427.37 & 5.33\% & 553.35 & -21.97\% & 1071.40 & -19.47\% \\
			& Distributional & 233.98 & 13.04\% & 335.64 & 9.29\% & 442.29 & 9.00\% & 621.80 & -12.32\% & 1077.95 & -18.98\% \\
			& Multi-step & 224.75 & 8.58\% & 319.07 & 3.89\% & 422.27 & 4.07\% & 550.50 & -22.37\% & 1037.15 & -22.04\% \\
			& Rainbow & 230.61 & 11.41\% & 324.35 & 5.61\% & 424.03 & 4.50\% & 559.90 & -21.05\% & 1057.50 & -20.51\% \\
			\cline{2-12}
			& Value (avg.) & \textbf{230.50} & \textbf{11.36\%} & \textbf{329.19} & \textbf{7.18\%} & \textbf{433.26} & \textbf{6.78\%} & \textbf{567.08} & \textbf{-20.03\%} & \textbf{1060.33} & \textbf{-20.30\%} \\
			\hline
			\multicolumn{2}{c|}{OR-Tools} & 1215.81 & - & 1751.77 & - & 2266.54 & - & 2989.30 & - & 5590.55 & - \\
			\hline
			\multirow{14}{*}{JSSP} & PPO & 1418.75 & 16.69\% & 1925.30 & 9.91\% & 2405.48 & 6.13\% & 3343.85 & 8.99\% & 5811.40 & 3.95\% \\
			& A2C & 1489.14 & 22.48\% & 2043.47 & 16.65\% & 2571.50 & 13.45\% & 3491.50 & 16.80\% & 6066.20 & 8.51\% \\
			& REINFORCE & 1453.88 & 19.58\% & 2041.95 & 16.62\% & 2564.18 & 13.13\% & 3491.75 & 16.81\% & 6164.50 & 10.27\% \\
			& V-MPO & 1481.74 & 21.87\% & 2017.04 & 15.14\% & 2528.07 & 11.54\% & 3622.90 & 21.20\% & 6281.40 & 12.36\% \\
			\cline{2-12}
			& Policy (avg.) & 1460.88 & 20.16\% & 2006.94 & 14.58\% & 2517.31 & 11.06\% & 3487.50 & 15.95\% & 6080.80 & 8.77\% \\
			\cline{2-12}
			& DQN & 1405.10 & 15.57\% & 1913.88 & 9.25\% & 2390.93 & 5.49\% & 3418.45 & 14.36\% & 5916.75 & 5.83\% \\
			& DDQN & 1437.87 & 18.26\% & 1960.86 & 11.94\% & 2422.76 & 6.89\% & 3330.95 & 11.43\% & 5827.55 & 4.24\% \\
			& PER & 1414.30 & 16.33\% & 1956.42 & 11.68\% & 2480.38 & 9.43\% & 3358.30 & 12.34\% & 5878.95 & 5.16\% \\
			& Dueling & 1430.15 & 17.63\% & 1940.17 & 10.75\% & 2469.42 & 8.95\% & 3374.70 & 12.89\% & 6048.40 & 8.19\% \\
			& Noisy & 1439.03 & 18.36\% & 1948.85 & 11.25\% & 2423.24 & 6.91\% & 3415.50 & 14.26\% & 5883.05 & 5.23\% \\
			& Distributional & 1500.23 & 23.29\% & 2123.65 & 21.23\% & 2708.16 & 19.48\% & 3678.00 & 23.04\% & 6090.10 & 8.94\% \\
			& Multi-step & 1496.04 & 23.05\% & 1997.89 & 14.05\% & 2491.88 & 9.94\% & 3508.20 & 17.36\% & 5927.10 & 6.02\% \\
			& Rainbow & 1444.36 & 18.80\% & 1974.20 & 12.70\% & 2453.36 & 8.24\% & 3555.65 & 18.95\% & 6162.80 & 10.24\% \\
			\cline{2-12}
			& Value (avg.) & \textbf{1445.97} & \textbf{18.91\%} & \textbf{1976.99} & \textbf{12.86\%} & \textbf{2480.02} & \textbf{9.42\%} & \textbf{3454.97} & \textbf{15.58\%} & \textbf{5966.84} & \textbf{6.73\%} \\
			\hline
		\end{tabular}
	}
\end{table*}

To complement the previous analysis, we next evaluate the models on a set of public benchmark datasets (Table~\ref{table_benchmark}), which include instances of varying sizes, derived from distributions that differ substantially from those used during training. This setup provides a strong foundation for assessing how well each model generalizes beyond their original training setting.

For these experiments, we employed a greedy decoding strategy with multiple starting nodes, following the approach proposed by \citet{Kwon2020}. In this decoding strategy, instead of generating a single solution per instance, multiple copies of the instance are created, one for each possible initial state. For the JSSP and FJSP, these initial states correspond to the set of initially assignable operation-machine pairs, i.e., the action space at $t=0$. Each copy is then solved independently using greedy decoding, starting with the specific initial action assigned to that copy. This strategy yields notable performance improvements during inference while incurring only negligible additional computational cost.

Across most tested datasets, we observed that, on average, value-based algorithms displayed superior generalization than policy-gradient ones. The only occasions where policy-gradient prevailed were on Demirkol and Dauzère and Paulli instances, with the latter showing a negligible difference between both categories of methods.

Among Rainbow extensions, the PER model displayed a very consistent performance across datasets, outperforming the DQN in 6 out of 7 datasets, indicating that, in terms of cross-size and cross-distribution generalization, it may be better than other algorithms. While other extensions generally outperformed PER on same-size, same-distribution instances (i.e., instances similar to the training data), they tended to perform worse on benchmark datasets, suggesting that some extensions may overfit to training data.

\begin{table*}[h]
	\caption{Results on public benchmark instances.}
	\label{table_benchmark}
	\centering
	\resizebox{\textwidth}{!}{
	\renewcommand{\arraystretch}{1.2}
		\begin{tabular}{c|cc|cc|cc|cc|cc|cc|cc}
			\hline
			\multirow{2}{*}{Method} & \multicolumn{2}{c|}{Brandimarte (FJSP)} & \multicolumn{2}{c|}{\makecell{Dauzère and Paulli\\ (FJSP)}} &
			\multicolumn{2}{c|}{\makecell{Chambers and Barnes\\ (FJSP)}} &
			\multicolumn{2}{c|}{\makecell{Hurink-vdata\\ (FJSP)}} & \multicolumn{2}{c|}{Taillard (JSSP)} &
			\multicolumn{2}{c|}{Demirkol (JSSP)} & \multicolumn{2}{c}{Lawrence (JSSP)} \\
			& $C_{max}$ & Gap & $C_{max}$ & Gap & $C_{max}$ & Gap & $C_{max}$ & Gap & $C_{max}$ & Gap & $C_{max}$ & Gap & $C_{max}$ & Gap \\
			\hline
			BKS & 297.00 & - & 2212.94 & - & 1008.43 & - & 920.00 & - & 2354.16 & - & 4614.23 & - & 1107.28 & - \\
			\hline
			PPO & 309.60 & 12.07\% & 2368.33 & 6.81\% & 1137.76 & 12.79\% & 935.80 & 1.86\% &  2662.79 & 15.04\% & 5636.73 & 21.74\% & 1199.78 & 8.62\% \\
			A2C & 311.40 & 13.69\% & 2417.78 & 9.00\% & 1137.76 & 12.84\% & 966.30 & 4.74\% & 2710.46 & 16.78\% & 5627.83 & 21.85\% & 1224.33 & 10.65\% \\
			REINFORCE & 321.20 & 15.50\% & 2407.00 & 8.47\% & 1153.95 & 14.58\% & 947.40 & 2.99\% & 2720.94 & 16.64\% & 5551.79 & 20.22\% & 1230.70 & 11.04\% \\
			V-MPO & 316.87 & 14.64\% & 2397.11 & 8.10\% & 1134.52 & 12.46\% & 937.80 & 2.15\% & 2830.76 & 21.92\% & 6081.61 & 30.81\% & 1232.15 & 11.58\% \\
			\cline{1-15}
			Policy (avg.) & 314.77 & 13.97\% & \textbf{2397.56} & \textbf{8.10\%} & 1141.00 & 13.17\% & 946.83 & 2.93\% & 2731.24 & 17.60\% & \textbf{5724.49} & \textbf{23.66\%} & 1221.74 & 10.47\% \\
			\cline{1-15}
			DQN & 313.33 & 13.62\% & 2422.50 & 9.22\% & 1123.76 & 11.39\% & 964.88 & 4.47\% & 2692.95 & 16.32\% & 5734.00 & 23.66\% & 1184.40 & 7.42\% \\
			DDQN & 314.40 & 13.75\% & 2385.94 & 7.58\% & 1139.71 & 13.09\% & 939.63 & 2.28\% & 2673.83 & 15.73\% & 5773.86 & 24.29\% & 1207.80 & 9.22\% \\
			PER & 311.67 & 12.99\% & 2386.17 & 7.57\% & 1116.67 & 10.80\% & 934.83 & 1.73\% & 2673.71 & 15.50\% & 5609.03 & 21.18\% & 1197.80 & 8.64\% \\
			Dueling & 315.33 & 13.74\% & 2371.67 & 6.91\% & 1132.19 & 12.30\% & 933.30 & 1.58\% & 2679.40 & 15.35\% & 5623.58 & 21.39\% & 1194.15 & 8.00\% \\
			Noisy & 313.87 & 13.51\% & 2404.56 & 8.41\% & 1142.90 & 13.49\% & 931.55 & 1.39\% & 2721.70 & 17.72\% & 5875.78 & 26.66\% & 1217.70 & 10.48\% \\
			Distributional & 310.13 & 13.20\% & 2387.89 & 7.72\% & 1142.48 & 13.40\% & 936.05 & 1.92\% & 2790.53 & 19.83\% & 6067.76 & 30.10\% & 1238.40 & 11.31\% \\
			Multi-step & 312.67 & 13.03\% & 2423.89 & 9.35\% & 1148.67 & 14.16\% & 937.03 & 2.06\% & 2739.49 & 18.60\% & 5923.85 & 27.62\% & 1225.98 & 10.67\% \\
			Rainbow & 308.40 & 12.51\% & 2399.50 & 8.11\% & 1154.33 & 14.52\% & 940.20 & 2.42\% & 2738.00 & 17.85\% & 5979.35 & 28.31\% & 1204.45 & 8.96\% \\
			\cline{1-15}
			Value (avg.) & \textbf{312.48} & \textbf{13.29\%} & 2397.77 & 8.11\% & \textbf{1137.59} & \textbf{12.89\%} & \textbf{939.68} & \textbf{2.23\%} & \textbf{2713.70} & \textbf{17.11\%} & 5823.40 & 25.40\% & \textbf{1208.83} & \textbf{9.34\%} \\
			\hline
		\end{tabular}
	}
\end{table*}

\subsection{Statistical validation}
\label{sec:stats}

\begin{figure*}[h]
	\centering
	\includegraphics[width=\textwidth]{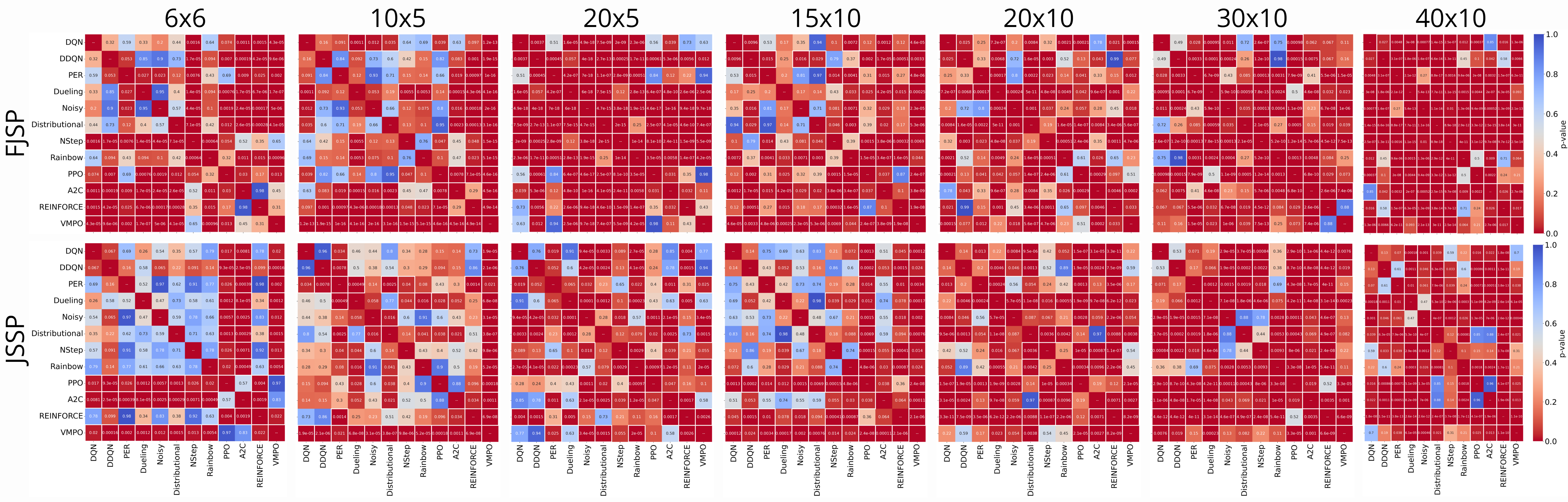}
	\caption{Heatmap showing the statistical differences between all tested algorithms, based on the Wilcoxon signed-rank test. The top row displays results for the FJSP, and the bottom row for the JSSP. Each column corresponds to a specific instance size. Statistically significant differences between algorithm pairs are marked with a white border around the corresponding cell.}
	\label{fig_stats}
\end{figure*}

To statistically validate our results, we conducted a Wilcoxon signed-rank test with a 95\% confidence level. The analysis was conducted independently for each problem type (JSSP and FJSP) and instance size, based on randomly generated instances. Each subplot on Figure~\ref{fig_stats} displays the pairwise test results between algorithms (rows vs. columns), indicating where statistically significant differences exist. Overall, the statistical findings were consistent with our earlier observations. In the FJSP, value-based methods were most often statistically better than policy-gradient algorithms. Among them, the Multi-step Learning and PER variants consistently achieved the strongest performance, while A2C was the most competitive among policy-gradient methods. For the JSSP, value-based algorithms showed significantly better performance on smaller instances, while policy-gradient methods performed better as the problem size increased. Within policy-gradient methods, REINFORCE and PPO were particularly strong, while the Distributional RL stood out among value-based algorithms.

These statistical findings reinforce the role of problem structure and characteristics in algorithm performance. The flexibility of the FJSP, with its two hierarchical decision levels, appears to favor value-based approaches. Additionally, value-based algorithms performed better on smaller problems, on both the JSSP and FJSP. 

\subsection{Runtime analysis}
\label{sec:runtimes}

\begin{figure*}[h]
	\centering
	\includegraphics[width=\textwidth]{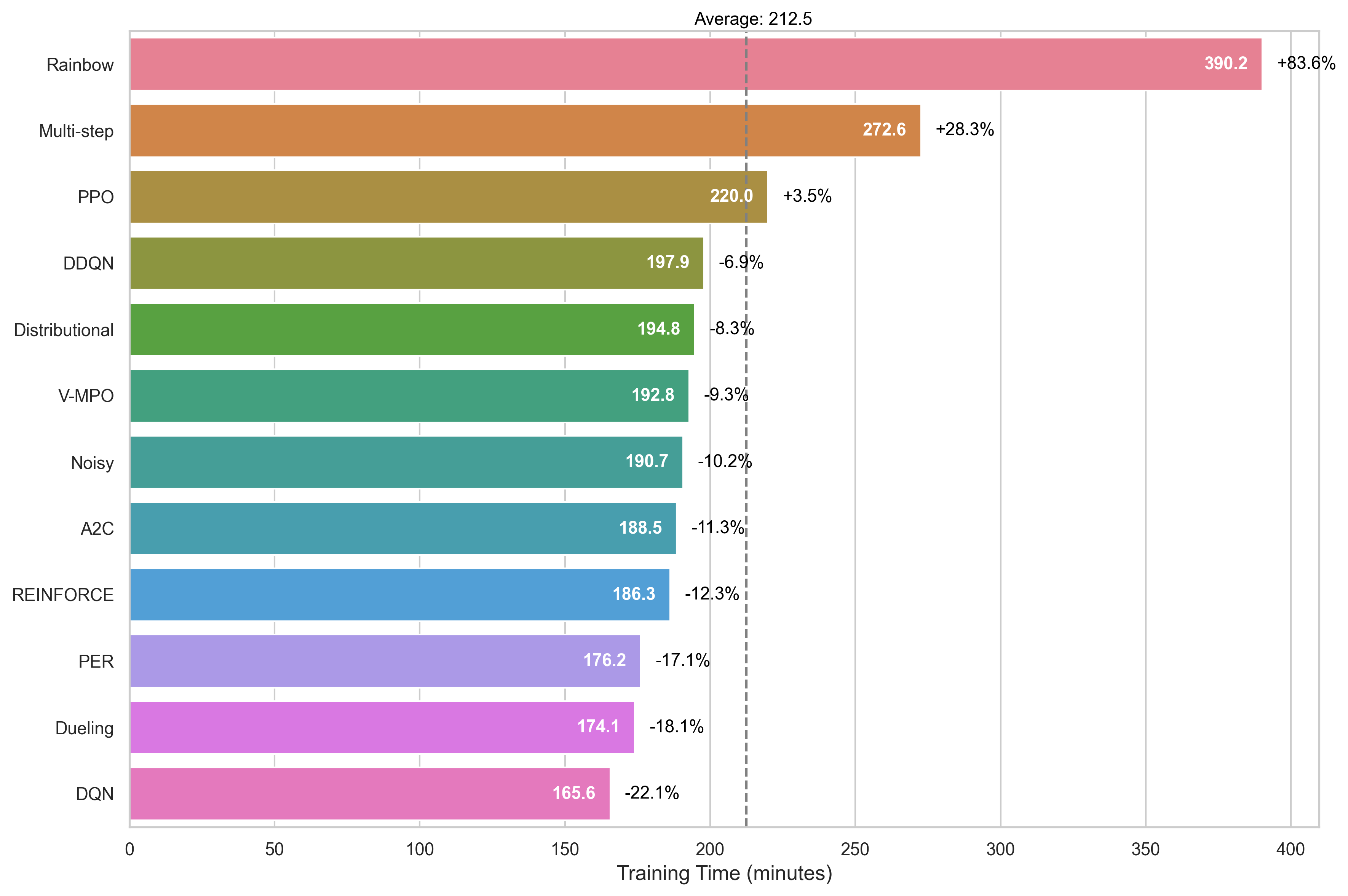}
	\caption{Average training runtime (minutes) for each RL algorithm.}
	\label{fig_training_runtime}
\end{figure*}

Although many value-based algorithms showed a generalization on par with policy-gradient methods, it is important to understand whether they also compete in terms of computational efficiency. Figure~\ref{fig_training_runtime} illustrates the average training runtime (in minutes) for each method evaluated in the previous sections. Additionally, we report the percentage deviation in training runtime relative to the average across all algorithms. Our results show that, in addition to being competitive in solution quality, value-based methods are also computationally efficient. With the exception of the full Rainbow model and the Multi-step learning extension, all other value-based methods trained faster than PPO and outperformed the average training runtime. The DQN, Dueling Network and PER algorithms stood out by achieving approximately a 20\% reduction in training runtime, with the DQN being the fastest algorithm out of all methods tested. Although the Noisy Network and Distributional RL models incurred slightly more computational cost, they remained competitive, with training times only marginally longer than other policy-gradient algorithms like REINFORCE and A2C.

We also report the average inference time (in seconds) taken by each algorithm, computed over the 100 testing instances of every instance size for both the JSSP and FJSP. The results are shown in Tables~\ref{table_inference_jssp} and \ref{table_inference_fjsp}. Overall, value-based algorithms achieved slightly lower inference times than policy-gradient methods across all instance sizes, on both the JSSP and FJSP.

\begin{table*}[!h]
	\caption{Average inference runtime (seconds) for each RL algorithm on 100 JSSP instances.}
	\label{table_inference_jssp}
	\centering
		\renewcommand{\arraystretch}{1.2}
		\begin{tabular}{c|cccccccc}
			\hline
			Algorithm & 6x6 & 10x5 & 20x5 & 15x10 & 20x10 & 30x10 & 40x10 & Average \\
			\hline
			PPO & 0.78 & 1.21 & 2.49 & 2.68 & 4.17 & 7.16 & 12.80 & 4.47 \\
			A2C & 0.86 & 1.06 & 2.35 & 2.61 & 4.23 & 6.88 & 12.75 & 4.39 \\
			REINFORCE & 0.92 & 1.07 & 2.00 & 2.82 & 4.09 & 6.80 & 12.73 & 4.35 \\
			V-MPO & 0.90 & 1.06 & 2.03 & 2.93 & 4.42 & 7.34 & 12.84 & 4.50 \\
			\hline
			Policy (avg.) & 0.86 & 1.10 & 2.22 & 2.76 & 4.23 & 7.04 & 12.78 & 4.43 \\
			\hline
			DQN & 0.63 & 0.77 & 1.65 & 1.95 & 3.10 & 5.86 & 11.03 & 3.57 \\
			DDQN & 0.60 & 0.72 & 1.66 & 1.84 & 2.85 & 5.65 & 11.12 & 3.51 \\
			PER & 0.66 & 0.75 & 1.45 & 1.84 & 3.07 & 6.03 & 10.99 & 3.55 \\
			Dueling & 0.62 & 0.82 & 1.52 & 1.93 & 3.06 & 6.05 & 11.56 & 3.68 \\
			Noisy & 0.67 & 0.76 & 1.58 & 1.87 & 3.19 & 6.31 & 11.32 & 3.68 \\
			Distributional & 0.57 & 0.77 & 1.44 & 1.69 & 3.01 & 6.08 & 11.29 & 3.58 \\
			Multi-step & 0.56 & 0.76 & 1.49 & 1.68 & 2.98 & 5.95 & 10.80 & 3.49 \\
			Rainbow & 0.64 & 0.92 & 1.60 & 1.84 & 3.20 & 6.52 & 12.18 & 3.88 \\
			\hline
			\textbf{Value (avg.)} & \textbf{0.62} & \textbf{0.78} & \textbf{1.55} & \textbf{1.83} & \textbf{3.06} & \textbf{6.06} & \textbf{11.29} & \textbf{3.61} \\
			\hline
	\end{tabular}
\end{table*}

\begin{table*}[!h]
	\caption{Average inference runtime (seconds) for each RL algorithm on 100 FJSP instances.}
	\label{table_inference_fjsp}
	\centering
		\renewcommand{\arraystretch}{1.2}
		\begin{tabular}{c|cccccccc}
			\hline
			Algorithm & 6x6 & 10x5 & 20x5 & 15x10 & 20x10 & 30x10 & 40x10 & Average \\
			\hline
			PPO & 0.60 & 0.76 & 1.74 & 2.92 & 4.11 & 6.87 & 12.95 & 4.28 \\
			A2C & 0.53 & 0.84 & 1.59 & 2.94 & 3.99 & 6.97 & 12.18 & 4.15 \\
			REINFORCE & 0.54 & 0.79 & 1.50 & 2.91 & 4.29 & 6.98 & 12.38 & 4.20 \\
			V-MPO & 0.59 & 0.85 & 1.86 & 2.63 & 4.22 & 7.23 & 13.07 & 4.35 \\
			\hline
			Policy (avg.) & 0.56 & 0.81 & 1.67 & 2.85 & 4.15 & 7.01 & 12.64 & 4.24 \\
			\hline
			DQN & 0.50 & 0.53 & 1.02 & 1.91 & 3.04 & 5.69 & 10.65 & 3.33 \\
			DDQN & 0.36 & 0.51 & 1.06 & 1.98 & 2.98 & 5.50 & 10.87 & 3.32 \\
			PER & 0.39 & 0.55 & 1.11 & 1.99 & 3.18 & 5.56 & 11.00 & 3.40 \\
			Dueling & 0.43 & 0.53 & 1.03 & 1.96 & 3.08 & 5.64 & 10.92 & 3.37 \\
			Noisy & 0.43 & 0.64 & 0.98 & 1.96 & 2.97 & 6.14 & 11.00 & 3.45 \\
			Distributional & 0.40 & 0.54 & 1.16 & 1.89 & 2.83 & 6.09 & 11.16 & 3.44 \\
			Multi-step & 0.36 & 0.48 & 1.14 & 1.90 & 2.96 & 6.18 & 11.22 & 3.46 \\
			Rainbow & 0.47 & 0.57 & 1.20 & 2.10 & 3.23 & 6.38 & 12.24 & 3.74 \\
			\hline
			\textbf{Value (avg.)} & \textbf{0.42} & \textbf{0.54} & \textbf{1.09} & \textbf{1.96} & \textbf{3.03} & \textbf{5.90} & \textbf{11.13} & \textbf{3.41} \\
			\hline
	\end{tabular}
\end{table*}

\section{Conclusion and future work}
\label{sec:conclusion}

In this work, we conducted an extensive evaluation of value-based RL algorithms on two canonical scheduling problems, demonstrating that these methods hold potential for solving complex CO problems. Our experiments revealed that value-based algorithms not only competed with, but in many cases outperformed several policy-gradient methods. These findings challenge the prevailing assumption that policy-gradient is inherently superior for CO, suggesting that value-based methods deserve more attention from the community. 

We observed that the lower variance of value-based algorithms held in a complex combinatorial domain, leading to a generally more stable convergence profile than policy-gradient methods. Interestingly, value-based algorithms also showed an impressive cross-size and cross-distribution generalization, being superior to policy-gradient across numerous public benchmark datasets.

Finally, we also found out that value-based and policy-gradient algorithms can have different effectiveness depending on specific properties of the problem. For instance, larger instances of the JSSP seemed to favor policy-gradient algorithms, while the FJSP, a problem with a more flexible structure, generally favors value-based. Among the value-based methods studied, we evaluated both the classic DQN and a set of advanced algorithmic extensions. While various extensions often improved upon the baseline DQN, their gains were generally modest, and not universally consistent across different problem types. As a result, replicating the effectiveness of these methods in complex combinatorial environments proved to be challenging, highlighting the need for further research to assess their applicability beyond classic RL benchmarks. Given the substantial implementation effort these extensions require, understanding their true practical value in real-world CO problems is crucial for decision-makers.

Looking ahead, we plan to further investigate the impact of problem complexity on algorithmic performance, extending our evaluation to richer and more realistic scheduling problems like the job-shop with setup times, machine breakdowns, and due dates.

\section*{Acknowledgements}

This work has been supported by the European Union under the Next Generation EU, through a grant of the Portuguese Republic's Recovery and Resilience Plan Partnership Agreement [project C645808870-00000067], within the scope of the project PRODUTECH R3 – "Agenda Mobilizadora da Fileira das Tecnologias de Produção para a Reindustrialização", Total project investment: 166.988.013,71 Euros; Total Grant: 97.111.730,27 Euros; and by the European Regional Development Fund (ERDF) through the Operational Program for Competitiveness and Internationalization (COMPETE 2020) under the project POCI-01-0247-FEDER-046102 (PRODUTECH4S\&C); and by national funds through FCT – Fundação para a Ciência e a Tecnologia, under projects UID/00285 - Centre for Mechanical Engineering, Materials and Processes and LA/P/0112/2020.

\bibliographystyle{elsarticle-harv} 
\bibliography{./bibliography}

\appendix

\section{Operation and machine node features}
\label{sec:appA}

In this appendix, we detail the different features used for each machine and operation node. For each operation node $O_{ij}$, there are six different features, which are:
\begin{itemize}
	\item \textit{Status}: A Boolean variable indicating whether the corresponding operation has already been scheduled or not.
	\item \textit{Number of neighboring machines}: The number of eligible machines currently connected to the operation, representing its flexibility in machine assignment.
	\item \textit{Processing time}: If the operation has been scheduled, this feature records the actual processing time on the assigned machine. Otherwise, it is set to the average processing time across all candidate machines.
	\item \textit{Starting time}: If the operation has been scheduled, this feature indicates its true starting time, i.e., the time in which the operation began to be processed. Otherwise, it represents an estimated starting time based on the current partial schedule $S(t)$, computed recursively using the operation precedence constraints. Specifically, for an unscheduled operation $O_{ij}$, the estimate is given by the starting time of its immediate predecessor $O_{i(j-1)}$, plus the processing time of $O_{i(j-1)}$ on its assigned machine. If $O_{i(j-1)}$ is also unscheduled, the estimate is calculated using the estimated starting time of $O_{i(j-1)}$, plus the average of its processing times across all eligible machines.
	\item \textit{Remaining operations in the job}: The number of unscheduled operations remaining in the same job as the current operation.
	\item \textit{Job completion time}: If all operations in $J_i$ have been scheduled, this is the actual job completion time $C_i$. Otherwise, this is the estimated completion time, based on the recursive calculations described above for the operations starting times.
\end{itemize}

For each machine node $M_k$, there are three different features:
\begin{itemize}
	\item \textit{Available time}: The earliest time at which $M_k$ becomes idle and ready to process a new operation.
	\item \textit{Number of neighboring operations}: The number of possible eligible operations to be assigned on $M_k$.
	\item \textit{Utilization rate}: The percentage of time $M_k$ has spent processing operations relative to the makespan of the current partial schedule.
\end{itemize}

\section{Hyperparameters}
\label{sec:appB}

The Rainbow algorithmic extensions significantly expand the hyperparameter search space compared to the original DQN algorithm. Consequently, performing an exhaustive hyperparameter search becomes impractical. In our study, we initially adopted the default values reported in the original papers for each extension. We then manually adjusted them based on their sensitivity, focusing on larger instances, which exhibited greater instability during training.

We set the exploration rate to start at $\varepsilon=1$, with an exponential decay of $e^{(-eps/600)}$, where $eps$ denotes the episode number. The exploration rate decreases until reaching a minimum value of 0.1, from which point it is maintained until the end of training. Across all scenarios, we used a learning rate of $2 \times 10^{-4}$ with the Adam optimizer. The minibatch size was set to 32, proving to be a lot more stable than using larger batches. The target network was updated every 10 episodes. We set the discount factor to 0.99, and the experience replay buffer capacity to 20000 transitions. For the PER, we set the prioritization exponent to 0.4. The importance sampling exponent began at 0.4 and increased linearly throughout training, reaching 1 by the end. In Multi-step learning, the update target was set to 4 steps for the FJSP and 2 for the JSSP. For the Distributional RL extension, we used 51 atoms. The minimum and maximum values used in the probability mass calculations were highly sensitive hyperparameters, occasionally causing instability during training. For JSSP instances, we set the minimum to -600 and the maximum to -50, whereas for FJSP instances, we used -50 and 0, respectively.

For all policy-gradient methods, we used a learning rate of $2 \times 10^{-4}$ with the Adam optimizer, and a discount factor $\gamma=1.0$. The loss function included a policy term weighted by 1.0 (in all methods besides V-MPO), a value term weighted by 0.5 (in all methods besides REINFORCE), and an entropy term with a coefficient of 0.01 to encourage exploration (in all methods besides V-MPO). For PPO, we used a clipping threshold of 0.2 and applied three gradient update steps per episode. Similarly, V-MPO was trained with three gradient update steps and both Lagrange multipliers --- $\alpha_{\text{V-MPO}}$ and $\eta_{\text{V-MPO}}$ --- were initialized at 1.0. The $\epsilon_{\eta \text{V-MPO}}$ constraint threshold was fixed at 0.01, while the $\epsilon_{\alpha \text{V-MPO}}$ constraint was sampled at each update from a log-uniform range between 0.001 and 0.01. Training was performed in parallel over multiple instances per episode: 20 for PPO and V-MPO, and 32 for A2C and REINFORCE. The validation step was performed every 10 episodes for all methods.

GNN-specific hyperparameters, that is, the number of GNN layers and hidden dimensions, were kept consistent with the original configuration \citep{Song2023} across all value-based and policy-gradient approaches.
	
\end{document}